\documentclass{article}
\usepackage{graphicx} 
\usepackage[natbibapa]{apacite}
\usepackage{subcaption}  

\usepackage{float}

\newfloat{textbox}{tbhp}{lop}
\floatname{textbox}{Box}

\usepackage{multirow}
\usepackage{multicol}
\usepackage{array}
\usepackage{xcolor}

\usepackage[framemethod=default]{mdframed}
\usepackage[a4paper, total={6in, 8in}]{geometry}

\usepackage{tcolorbox}
\usepackage{caption}

\usepackage[toc,page]{appendix} 
\newtcolorbox{textexcerpt}[2][]{%
  colback=white,
  colframe=black,
  fonttitle=\bfseries,
  #1
}

\usepackage{orcidlink}

\usepackage{authblk}

\usepackage{amsmath}

\usepackage{float}

\usepackage{listings}
\definecolor{codegreen}{rgb}{0,0.6,0}
\definecolor{codegray}{rgb}{0.5,0.5,0.5}
\definecolor{codepurple}{rgb}{0.58,0,0.82}
\definecolor{backcolour}{rgb}{0.95,0.95,0.92}

\lstdefinestyle{mystyle}{
    backgroundcolor=\color{backcolour},   
    commentstyle=\color{codegreen},
    keywordstyle=\color{magenta},
    numberstyle=\tiny\color{codegray},
    stringstyle=\color{codepurple},
    basicstyle=\ttfamily\footnotesize,
    breakatwhitespace=false,         
    breaklines=true,                 
    captionpos=b,                    
    keepspaces=true,                 
    numbers=left,                    
    numbersep=5pt,                  
    showspaces=false,                
    showstringspaces=false,
    showtabs=false,                  
    tabsize=2
}

\title{Unlocking the Potential of Past Research: Using Generative AI to Reconstruct Healthcare Simulation Models}

\author{
Thomas Monks \textsuperscript{a}\orcidlink{0000-0003-2631-4481}\thanks{CONTACT Thomas Monks. Email: t.m.w.monks@exeter.ac.uk},
Alison Harper, \textsuperscript{b}\orcidlink{0000-0001-5274-5037},
Amy Heather \textsuperscript{a}\orcidlink{0000-0002-6596-3479}}

\affil{\textsuperscript{a}University of Exeter Medical School, St Luke's Campus, Heavitree Rd, Exeter, UK;
\textsuperscript{b}Centre for Simulation, Analytics and Modelling (CSAM), University of Exeter Business School, Streatham Campus, Rennes Drive, Exeter, UK.
}

\begin{document}

\maketitle

\begin{abstract}

\noindent Discrete-event simulation (DES) is widely used in healthcare Operations Research, but the models themselves are rarely shared. This limits their potential for reuse and long-term impact in the modelling and healthcare communities. This study explores the feasibility of using generative artificial intelligence (AI) to recreate published models using Free and Open Source Software (FOSS), based on the descriptions provided in an academic journal. Using a structured methodology, we successfully generated, tested and internally reproduced two DES models, including user interfaces. The reported results were replicated for one model, but not the other, likely due to missing information on distributions. These models are substantially more complex than AI-generated DES models published to date. Given the challenges we faced in prompt engineering, code generation, and model testing, we conclude that our iterative approach to model development, systematic comparison and testing, and the expertise of our team were necessary to the success of our recreated simulation models.   

\end{abstract}


\section{Introduction}

Operational Research (OR) has a long history of working alongside health services to use models to support decisions aiming to improve patient health outcomes and reduce service costs. Discrete-Event Simulation (DES) is the most widely applied simulation method. Given its maturity, there are now multiple reviews of DES in health \citep{soorapanth2023towards,liu2020diffusion,vazquez2021discrete,zhang2018application,forbus2022discrete} that highlight the breadth of DES application in areas that include chronic lung disease, COVID-19 planning, cancer services, neonatal pathways, emergency care, and stroke treatment services \citep{harper2023strategic,gjerloev2024systematic,penn2024infant,salmon2018structured, ouda2023comprehensive,koca2024systematic,yakutcan2022patient,campos2023integrating}.

At the time of writing, around 100 DES articles applied to health and medical problems are published in scholarly journals, conferences and books each year \citep{monks2024towards}. Many of the articles reporting applied research studies directly worked with a health service to support real decisions. In each case, key outputs of the research are the conceptual and coded DES models. The conceptual model might take many forms in the report: from an unstructured narrative description to a more formal description, such as following a conceptual modelling framework \citep{robinson2008conceptual2} or as prescribed by a reporting guideline \citep{stress_guidelines,grimm2020odd}. Whatever the approach used by the study authors, the conceptual models are published - they are \textit{available} to interested parties to reuse if they choose.  On the other hand, the coded model - the digital artefact that generates the results reported in the paper - is rarely made available in any form - as low as 8\% \citep{monks2024towards}.  The coded model remains with the original authors, but over time, as authors move on, or retire, it is effectively lost to the modelling and simulation community. We note that the development of the coded model is expensive. Not only did it require modeller time to code and analyse, but it also required time from medical professionals, healthcare managers, informatics specialists, and service users (patients and their families) to provide domain expertise, support validation, and challenge assumptions or results.  This cost could be voluntary or could have been supported by public funding.

The publication of coded models alongside a conceptual model falls under the field of Open Science. The Facilitating Open Science in European Research (FOSTER) project\footnote{\url{https://book.fosteropenscience.eu/}} \citep{bezjak2018handbook} defines Open Science as ``the practice of science in such a way that others can collaborate and contribute, where research data, lab notes and other research processes are freely available, under terms that enable reuse, redistribution and reproduction of the research and its underlying data and methods". In this article we focus on the reuse potential of DES models of health services that can be realised through Open Science.  While the availability of conceptual models supports reuse, the cost of re-coding a DES model from scratch is likely prohibitive. This is particularly true from the perspective of a health service reuser. While an academic DES study will typically run over months or years, a health service may need a model to use within weeks or potentially days.  We argue that the lack of availability of coded DES models reduces the potential long-term impact of DES research in health services.  

Since the release of ChatGPT\footnote{\url{https://chatgpt.com/}} there has been excitement about the potential of Artificial Intelligence (AI) to increase productivity. Pioneering simulation research proposes a ``Natural Language Processing (NLP) Shortcut" framework \citep{jackson_2024} for generating DES models. This framework combines the general coding abilities of generative AI and Large Language Models (LLMs) with prompt engineering to create a coded \textit{and reusable} simulation model. 

To date, generative AI has only been used to code simple simulation models; for example consisting of 20-30 lines of Python code \citep{jackson_2024,frydenlund2024modeler} or to support a reflective modelling process \citep{Giabbanelli_GPT_Sim}. The level of detail and scope of these models are not reflective of DES models used in practice within health services. Such a model would, for example, require hundreds of lines of code, include more sophisticated process logic (e.g., balking, probabilistic routing, and multiple entity classes), make use of coding best practices, control pseudorandom number streams, and include a user interface. Our aim in this study is to build on the pioneering NLP Shortcut framework to investigate the potential of generative AI to recreate coded models from the conceptual model of a healthcare DES model reported in academic publications.  We note that many DES studies make use of Commercial Off-the Shelf (COTS) packages for their simulation modelling. We take an Open Science approach to recreating these models and generate Python code. This code is then licensed and preserved using Open Science archives and a research compendium \citep{gentleman_statistical_2007,ayllon_keeping_2021}.

\section{Aims}

Our study investigates the feasibility of using generative AI to recreate DES models in healthcare based on textual descriptions from the academic literature. We focus on generating models in the Python simulation package \textit{SimPy} \citep{simpy}, selected for its (i) compatibility with language models' code-generating capabilities, (ii) growing adoption in health service Operational Research \citep{monks_harper_des_review}, and (iii) our expertise in developing \textit{SimPy} models for healthcare applications \citep{harper2023development,allen2020simulation}.

To assess feasibility, we engineer prompts for Perplexity.AI to generate complete Python and \textit{SimPy} code that captures model logic (e.g. arrival processes, queuing, activities, sampling, and balking). Additionally, we explore generating browser-based user interfaces using \textit{Streamlit} \citep{streamlit} to enhance accessibility for non-programmers. Our research objectives are to:

\begin{itemize}
\item Determine if generative AI can produce functional, verifiable \textit{SimPy} models from engineered prompts describing DES models
\item Assess the feasibility of generating usable \textit{Streamlit} web interfaces for these models
\item Pilot this approach by recreating two published healthcare DES models
\item Evaluate the reproducibility of our methodology when conducted by different modellers
\end{itemize}

This work contributes to the growing interest in generative AI applications for modeling \citep{tolk2024hybrid,frydenlund2024modeler,giabbanelli2024broadening,Giabbanelli_GPT_Sim}. Our long-term goal is to develop guidance on prompt engineering and to document the opportunities, challenges, and limitations of using AI to recreate DES models—ultimately supporting result reproduction, model reuse, and educational applications.

The remainder of this paper first defines key concepts in generative AI, LLMs, and chatbots, followed by a review of generative AI applications in simulation modelling. We then detail our methodology and apply it to two case studies: a 2010 critical care model and a 2016 stroke pathway model. Finally, we present our results and discuss implications for Operational Research in this emerging field.
\section{Generative AI}

Before reviewing relevant generative AI research for simulation, we briefly define generative AI and describe popular LLMs and human interaction with them via Chatbot AI tools. We summarise the key concepts in Table \ref{tab:concepts}.

\begin{table}
\footnotesize
\centering
\caption{Key Concepts in Generative AI.}
\label{tab:concepts}
\begin{tabular}{|p{5cm}|p{7cm}|}
\hline
\multicolumn{1}{|c|}{\textbf{Topic}}                       & \multicolumn{1}{c|}{\textbf{Summary}}                                                                                                  \\ \hline
Generative AI                                     & AI models designed to create novel digital content such as text, images, music, or code.                                      \\ \hline
Large Language Models (LLMs)                      & A subset of generative AI specialising in processing and generating human-like text.                                          \\ \hline
Transformer Architecture                          & Neural network design using self-attention mechanisms to process and generate text.                                           \\ \hline
Zero-Shot Learning                                & The ability of a model to perform tasks or make predictions on categories it hasn’t explicitly seen during training.          \\ \hline
Model Scaling                                     & The process of increasing model size (number of parameters) to improve performance and capabilities.                          \\ \hline
Hallucination                                     & The tendency of LLMs to generate plausible-sounding but factually incorrect or logically flawed content.                      \\ \hline
Data Contamination                                & The overlap of training data with test data, potentially leading to overestimated model performance.                          \\ \hline
Temperature                                       & A parameter controlling the randomness and creativity in LLM outputs.                                                         \\ \hline
Prompt Engineering                                & The process of crafting effective inputs to elicit desired outputs from LLMs.                                                 \\ \hline
Chatbot AI                                        & AI-powered conversational interfaces that use LLMs to understand and generate human-like responses in real-time interactions. \\ \hline
Context Window                                    & The amount of previous conversation an LLM can consider when generating responses.                                            \\ \hline
RLHF (Reinforcement Learning from Human Feedback) & A technique used to fine-tune LLMs based on human ratings of model outputs.                                                   \\ \hline
Alignment Problem                                 & The challenge of ensuring AI outputs align with human values and intentions.                                                  \\ \hline

Retrieval Augmented Generation                                 & A technique that enhances LLMs by integrating an external information retrieval system.                                                  \\ \hline

\end{tabular}
\end{table}

\subsection{Generating novel content using LLMs}

Traditional Machine Learning (ML) paradigms, such as classification, train a model to learn patterns within historical labelled data in order to classify new unseen instances. For example, classifying if a brain scan indicates Parkinson's Disease or is healthy. Generative AI models are trained on unlabelled data and, rather than predict or classify, they aim to create novel digital content such as text, images, music, or code. One example is generating a simple simulation model in Python code \citep{jackson_2024,frydenlund2024modeler}. LLMs are a subset of generative AI that specialise in natural language communication between humans and computers. The Generative Pre-trained Transformer (GPT) architecture, that underpins AI Chatbot tools like ChatGPT, is perhaps the most well-known example of an LLM. GPT models are built on transformer-based neural network architectures, which use self-attention mechanisms to process and generate text \citep{brown2020languagemodelsfewshotlearners,vaswani2023attentionneed}. In simple terms, GPT models are sequence predictors, trained to predict the next token (e.g., a word) in a sequence based on the context of previous tokens. 

\subsection{Zero-shot learning and model scaling}

A key advancement that distinguishes LLMs from traditional ML approaches is their capacity for zero-shot learning - the ability to perform tasks on previously unseen categories without explicit training \citep{brown2020languagemodelsfewshotlearners}. This capability enables LLMs to adapt to novel contexts and tasks, such as generating code based on user specifications, without additional training. The evolution of zero-shot learning has been closely tied to the increasing scale of language models. When GPT-1 was introduced in 2018, it contained 117 million parameters \citep{Radford2018ImprovingLU}. Subsequent iterations have seen substantial growth in model size, with GPT-3 including 175 billion parameters \citep{brown2020languagemodelsfewshotlearners}. The exact specifications of GPT-4 and GPT-4.5 have not been officially confirmed by OpenAI, but GPT-4 is speculated to contain up to a trillion parameters \citep{Giabbanelli_GPT_Sim}. 

\subsection{Challenges and limitations: data contamination and hallucination}

Evaluating the zero-shot capabilities of LLMs is challenging due to the potential contamination of test data \citep{xu2024benchmarkdatacontaminationlarge}. The concept of contamination is analogous to leakage in traditional supervised machine learning \citep{leakage_reference}, i.e., the training data overlaps with test data, accuracy measures are overstated, and the model is simply outputting data it has memorised in training. In the case of LLMs, it is difficult to determine if the training data overlaps with test data and careful evaluations must be designed.

A key challenge in the use of LLMs is mitigating the risk of \textit{hallucination}. LLMs are sequence prediction models that prioritise generating the most probable next word in a sequence, even if it is inaccurate. Simply put, given an input, a model will always produce an output, whether or not it is correct. As a result, an LLM may ``hallucinate": confidently present content that is factually incorrect, logically flawed, or at odds with the training data provided \citep{huang2023surveyhallucinationlargelanguage,Ziwei_hallucinations_ref2,dou2024s}.

For example, an LLM might generate plausible but fabricated references in an academic essay or produce code that appears functional but contains logical errors. These errors may go unnoticed by users, and have consequences that vary from minor (e.g., wasted time from debugging nonsensical code) to severe (e.g., incorrect decisions based on the results of a flawed simulation model). The causes of hallucination are complex and varied. In coding, for instance, it might stem from pre-training the LLM on code that contains both obvious and subtle bugs.

Hallucination is a major limitation of generative AI and hence is an active area of research \citep{Ziwei_hallucinations_ref2}. Promising approaches include variations on the theme of iterative retrieval of information \citep{khot2023_iterative,yao2023_iterative} that can involve refining outputs through multiple iterations each providing more context or fact-checking. Another approach is to estimate model uncertainty statistics that can highlight LLM knowledge deficiencies \citep{Farquhar2024}. For the immediate future, it seems likely that hallucination will continue to be a major challenge for the safe and productive use of generative AI, with some arguing it cannot be fully eliminated \citep{xu2024_hallucination_inevitable}. As such, it is crucial to incorporate some form of fact-checking or testing mechanisms in any work that relies on content generated by an LLM.

\subsection{Randomness and prompt engineering}

LLMs include an element of randomness in the generation of responses.  This randomness is typically controlled by a ``temperature" parameter, where higher values increase variability in outputs (and increase hallucinations), while lower values produce more deterministic results. The use of randomness allows LLMs to generate diverse and creative solutions but it also means that, given the same prompt, an LLM may produce different code outputs across multiple runs. This variability poses challenges for reproducibility in contexts such as code generation for simulation models, where consistent and replicable results are important. By default, Chatbot AI tools may not offer direct user control over temperature.

Given the randomness used in generative AI, and as LLMs tend to hallucinate, another important concept to define is the formation of prompts. This has given rise to the discipline of prompt engineering: the process of writing a prompt that results in the most effective LLM performance \citep{liu2021pretrainpromptpredictsystematic,pornprasit2024fine}. This is a very recent area of research and there is not yet a consensus on the most effective approaches although various patterns are available \citep{white2023promptpatterncatalogenhance,wang2024enhancing}. For example, in \textit{one-shot }or \textit{few-shot} learning, the prompt includes one or more simple examples of the task to clarify the context for the LLM.

\subsection{Retrieval Augmented Generation}

Retrieval Augmented Generation (RAG) is a promising research area that can reduce the occurrence of hallucinations in language models \citep{shuster2021_RAGreduceshallucination}. In the RAG process, a knowledge base is queried, and relevant information is incorporated into the prompt's context before it is processed by the language model \citep{lewis2021_RAG_intro}. In simple terms, this can be thought of as providing the language model with more factual information within the prompt. A knowledge base could be the internet; for example, Perplexity.AI retrieves web pages, and online documents prior to generating an answer.  Hallucinations can be reduced as RAG provides trained LLMs access to knowledge it has not seen before; for example, up to date information not used in training, or advancements in scientific fields that came post training. Hallucinations are not completely mitigated by RAG, for example, if a user's question cannot be answered from the knowledge base then the language model may still generate an incorrect answer \citep{barnett2024_7_Fails_RAG}.

\subsection{AI Chatbots and alignment}

Since 2022, and at the time of writing, wide-scale public access to LLMs has been made possible by general-purpose Chatbot AI tools such as Open AI's ChatGPT\footnote{\url{https://chatgpt.com/}}, Perplexity.AI's Sonar\footnote{\url{http://perplexity.ai/}}, Anthropic's Claude \footnote{\url{https://claude.ai}}, DeepSeek-R1 \citep{deepseekR1_paper} and Google's Gemini\footnote{\url{https://gemini.google.com/}}. The underlying LLMs are trained on large amounts of curated web data (including code from sources such as StackOverFlow and GitHub) and fine-tuned for chat-based human interaction. In general, the tools have been shown to understand and generate human-like text (and code) across a wide range of tasks. The overall architecture and training of these models is complex and for most models is not fully known given the commercial nature of the companies that create and operate them at substantial cost. As a general rule, however, LLMs such as GPT-3.5 or 4 are not used as is; instead, a new round of training is undertaken using a curated question and answer dataset. This process of fine tuning produces an Assistant Model that provides chat like responses to a human prompt. A further round of training employs reinforcement learning from human feedback (RLHF) where a human workforce reviews and rates responses output by the model \citep{casper2023RLHFlimitations}. RLHF aims to help Chatbot AI's tools align responses with the human values and the intentions of their prompts (the so-called ``alignment problem"). This process attempts to filter out inappropriate or offensive content while enhancing the models' ability to provide a relevant response. 

Human interaction with these models is via a user-friendly chat interface. The underpinning LLM in use varies by free and paid tiers (e.g., at the time of writing ChatGPT offers a free GPT-3.5 or paid GPT-4/4.5 tier). While the LLM architectures have no memory of prior prompts, a Chatbot AI tool has a context window allowing a user to interact iteratively with an LLM within a larger history/context of prompts and responses. There are size restrictions on these context windows that vary with each Chatbot AI tool and underlying model.

\subsection{Generative AI in software engineering}

LLMs have also been fine-tuned for software engineering and coding tasks such as code generation and code completion. Contemporary tools include GitHub Copilot\footnote{\url{https://copilot.github.com}} and Meta Code Llama \footnote{\url{https://llama.meta.com/code-llama}}.  Research in this area has been extensive with the vast majority spread across software development and maintenance\citep{hou2024largelanguagemodelssoftware}.

Code generation has used a mix of general LLM tools such as GPT-3.5 / 4.0 \citep{yetiştiren2023evaluatingcodequalityaiassisted,dou2024s} and fine-tuned models such as GPT-3's codex \citep{chen2021evaluatinglargelanguagemodels}. Performance of LLMs in code generation in these studies typically makes use of a curated benchmark dataset of programming problems and their solutions such as the \textit{Mostly Basic Python Problems Dataset}\footnote{\url{https://github.com/google-research/google-research/tree/master/mbpp}} \citep{austin2021programsynthesislargelanguage}. LLM solution performance against these datasets are evaluated using various standard metrics.

Prompt engineering to reduce ambiguity of intent has emerged as a key challenge for code generation as LLMs, which may struggle to reliably generate code reflecting the users' requirements. To date the majority of studies have employed zero- or few-shot prompt engineering to maximise LLM effectiveness \citep{hou2024largelanguagemodelssoftware}; a smaller number have explored novel approaches such as prompting LLMs to include a planning phase before generating solutions \citep{jiang2024selfplanningcodegenerationlarge}.

\subsection{Generative AI and computer simulation}

\subsubsection{Automated code generation}

Recent research has explored the integration of generative AI with computer simulation, yielding promising hybrid approaches. Several pioneering studies have investigated small-scale applications and conceptual frameworks across DES, system dynamics, conceptual modeling, and documentation \citep{jackson_2024,Akhavan_2024,Shrestha_gpt_explain_model,Giabbanelli_GPT_Sim,plooy_ai_2023}.

\cite{jackson_2024} explored the potential of using GPT-based models to produce simulation models for inventory and process control in logistics systems. Their research focused on the concept of an ``NLP Shortcut", where simulation models are generated from a textual description of the model passed to a language model. The study used the OpenAI Davinci Codex (a code-based Application Programming Interface (API) to the GPT-3 model) to successfully generate simple Python-based simulations of logistics systems (e.g. a single-product inventory-control system). The LLM outputs consist of 20-30 lines of Python code implementing simple DES model logic and code to plot the model output. Their framework incorporated dynamic execution of the generated code with human expert oversight, demonstrating the potential for AI-assisted simulation development.

\cite{Akhavan_2024} and \cite{plooy_ai_2023} investigated the application of ChatGPT in system dynamics modelling. Both studies take the position that generative AI should not replace a modeller but rather serve as a tool to facilitate the research process, review content, and enhance idea implementation in simulation modelling. \cite{Akhavan_2024} develop a simple System Dynamics model of COVID-19's impact on economic growth. Their approach first prompts ChatGPT (GPT-4) in an iterative manner to support conceptual modelling and decisions about methods. The authors \textit{manually code} a small Python model (40 lines of code) and provide this along with prompts to ChatGPT to generate suggestions for code optimisations, additional plotting code, and improvements to model documentation.

\cite{plooy_ai_2023} focused on using ChatGPT (GPT-4) to generate Python code implementing a simple system dynamics model of a resource-bound population in equilibrium. They outline a six-step approach to iteratively generate a model with ChatGPT's help. Early steps focus on textual information describing equations for stocks and flows that are first manually implemented in the commercial simulation package iSee Stella Architect\footnote{https://iseesystems.com}. The final step converts the generated equations into 32 lines of Python code with outputs verified by comparing the manually created and generated models.

\subsubsection{Conceptual modelling}

\cite{Giabbanelli_GPT_Sim} published a conceptual study that hypothesised the potential of LLM application across common simulation tasks. The study focused on four key areas: structuring conceptual models, summarising simulation outputs, improving accessibility to simulation platforms, and explaining simulation errors with guidance for resolution. For example, the potential to use the emerging capability of LLMs to convert images to text to provide automated explanations of charts of simulation output could benefit both non-experts and people with visual impairments.

\cite{Shrestha_gpt_explain_model} proposed a method where generative AI is used to explain simulation models, by creating simple conceptual model descriptions from more complex causal maps. Their approach involved decomposing large conceptual models into smaller parts and then performing Natural Language Generation (NLG) using a fine-tuned GPT-3 model. 

\subsection{Summary of lessons from the literature}

The application of generative AI to practical domains, including computer simulation modeling, is an emerging and rapidly evolving field of research. Our study has been designed to account for the following lessons and challenges identified in the existing literature:

\begin{itemize}
    \item \textbf{User Expertise}: The effectiveness of generative AI can depend on the expertise and skill level of the user.
    \item \textbf{Mitigating Hallucination}: Generative AI models are prone to hallucination (producing incorrect or fabricated outputs). This risk can be reduced through strategies such as prompt engineering, retrieval-augmented generation (RAG), and iterative refinement during interactions.
    \item \textbf{Model Validation}: AI-generated models require thorough testing and validation at each stage of development to ensure reliability and accuracy.
    \item \textbf{Selection of Test Data}: The choice of test data is critical to avoid data leakage, which can compromise the validity of results in generative AI studies.
    \item \textbf{Model Complexity}: Current research has predominantly focused on applying generative AI to relatively simple models, leaving its performance with more complex systems largely unexplored.
\end{itemize}

\section{Methods overview}

\begin{figure}
    \centering
    \includegraphics[width=1\linewidth]{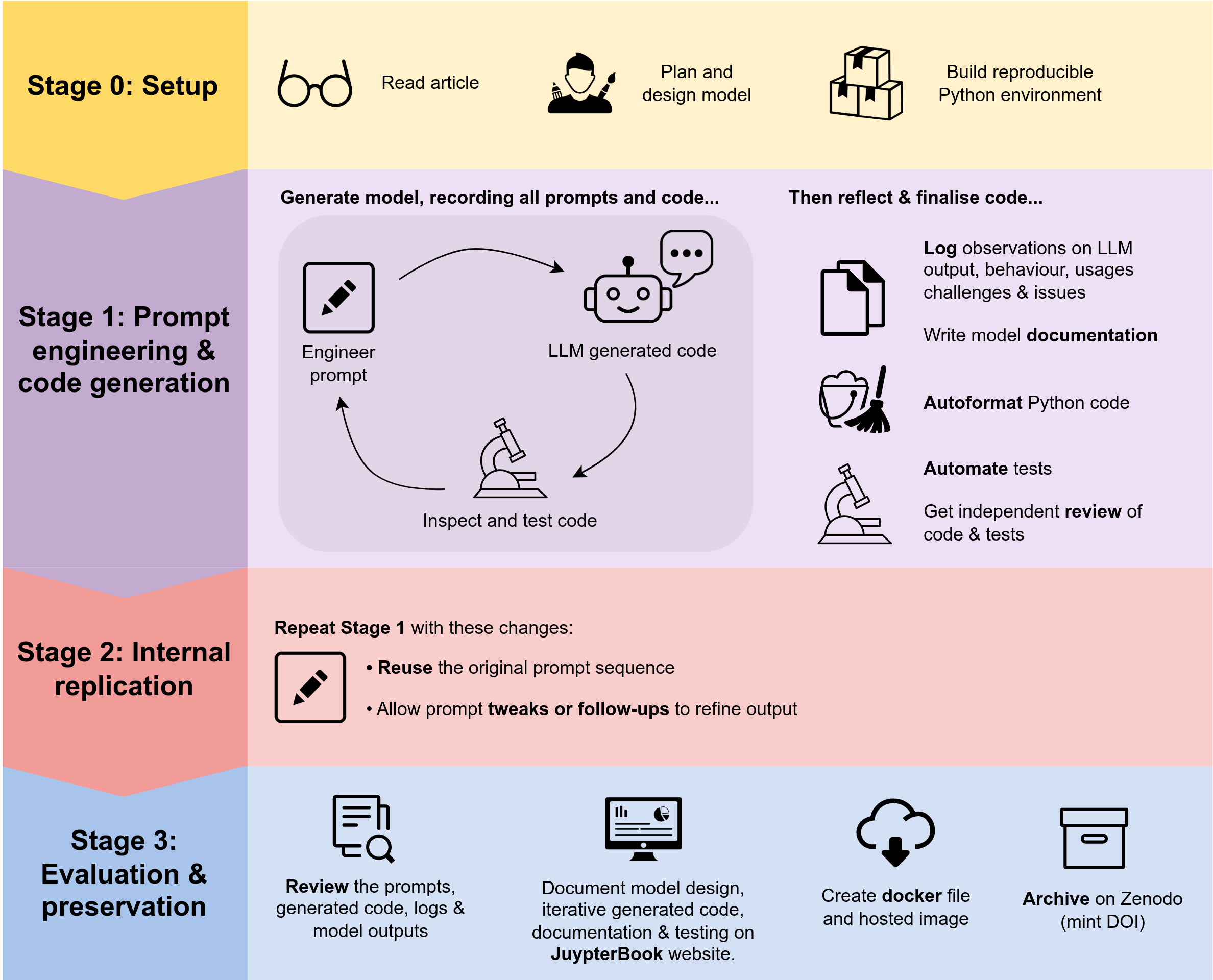}
    \caption{Overview of approach to model recreation using generative AI. \\ Abbreviations: AI, artificial intelligence; DOI, digital object identifier; LLM, large language model.}
    \label{fig:experiment_fig}
\end{figure}

Our study followed four stages: setup and model design (Stage 0); prompt engineering and code generation (Stage 1); internal replication (Stage 2); and evaluation and preservation (Stage 3). Figure \ref{fig:experiment_fig} illustrates these stages and the activities carried out in each. For model generation, we used Perplexity.AI's standard model (free tier) that includes RAG from internet sources. The RAG functionality provides the model with up-to-date and new online sources about simulation and \textit{SimPy}. 

\subsection{Stage 0: setup and model design}

For the two chosen case studies, we read the academic paper and wrote down a design for a simulation model organised by the \textit{Strengthening The Reporting of Empirical Simulation Studies} (STRESS) guideline format for reporting models \citep{stress_guidelines}. Previous studies of replicating models from journal articles have been challenging due to reporting ambiguities \citep{schwander_replication_2021,mcmanus2019can}.  We therefore document any simplifications or additional assumptions (e.g., undocumented parameters or logic or removal of a feature) that were made to enable us to design a functioning version of the model reported in the journal article. We designed a common Python 3.10 software environment (implemented as a conda\footnote{\url{https://anaconda.org/anaconda/conda}} environment) that could be used in stages 1 to 3 to run the generated code.

\subsection{Stage 1: prompt engineering and code generation}

In Stage 1, we created a prompt database that stored all prompts given to the LLM and for what purpose. We then proceeded to generate the design of the model as reported in the paper for Case Study 1. When complete, we recreated Case Study 2. Where appropriate we reused or adapted prompts from Case Study 1 when recreating the model for Case Study 2. All case study prompts were stored in the database in sequence of use. After the models were complete we created a research compendium consisting of a STRESS report for the design, formatted model code (using the tool `black'), a script to run the model, documentation for the user interface, and an automated model test suite. A second modeller then reviewed the research compendium and identified any Stage 1 modifications required before proceeding to Stage 2.

\subsection{Stage 2: internal reproduction}

In Stage 2, a second modeller reproduced the process from Stage 1 by attempting to recreate the models using the original sequence of prompts. Given the stochastic nature of creative LLM output (i.e. the same input prompt may not produce the same output), we allowed for the re-engineering of prompts in the replication phase. This included direct modification of the original prompts and additional follow-up prompts to refine the output. We also allowed for structural differences between the original and replicated models, for example, the use of different Python data structures, and class/function designs. Stage 2 was added to the research compendium and independently checked in an identical manner to Stage 1.

\subsection{Stage 3: evaluation and preservation}

\subsubsection{Evaluation}
In Stage 3, we compared the artefacts generated and experience of working with the LLM to create the \textit{SimPy} models in Stages 1 and 2.  We defined a successful internal replication to be when Stages 1 and 2 models produced the same results. As we designed the models to use the same seeds and random number generators we aimed for identical results; however, we allowed a small tolerance of 5\% in line with other replication studies \citep{mcmanus2019can,schwander_replication_2021}. To evaluate the use of the LLM for generating the models, the modellers from Stages 1 and 2 synthesised their experience of prompting the LLM: identifying common successes/failings, general challenges, coding mistakes, and opportunities.

\subsubsection{Model Preservation}
The final step in Stage 3 was to preserve the models we had recreated, ensuring that they are available to others to inspect or use long term. We structured all of our materials into a research compendium - a website compiled using the tool JupyterBook\footnote{https://jupyterbook.org}. The compendium consists of the code, data, and documentation needed to understand and reproduce our study. We then deposited the research compendium in the Zenodo open science archive\footnote{\url{http://zenodo.org/}} and obtained a Digital Object Identifier (DOI).

\section{Case study selection}

We selected two published healthcare DES case studies. Prior LLM simulation coding studies had focused on very simple coding tasks comprising 20-30 lines of code. We aimed to increase the complexity of the coding task for the LLM in our study. We selected DES models that consisted of multiple classes of patients (e.g., multiple arrival sources and differing sampling distributions for length of stay), and at least two activities (delays). From experience, we estimated that design of such models would require between between 200 - 500 lines of Python code. We would still classify these as simple DES models.

A threat to the external validity of our study is selecting a DES study where the exact or a very similar model is available within the training data of the LLM.  The risk is that the LLM simply outputs the training data when prompted. The exact training data of commercial LLMs is unknown, but we assumed it would include popular code repositories such as GitHub\footnote{\url{http://github.com/}} and coding question-and-answer sites such as Stack Overflow.\footnote{\url{https://stackoverflow.com/}} As we are using Python, one way to reduce this risk is to select a DES study where the model has been reported to be developed in a commercial simulation package interface such as Simul8,\footnote{\url{https://www.simul8.com/}} Arena,\footnote{\url{https://www.rockwellautomation.com/en-gb/products/software/arena-simulation.html}} or Excel.\footnote{\url{https://www.microsoft.com/en-gb/microsoft-365/excel}}  The majority of such models are contained within a bespoke proprietary format and not the natural (or coding) language that a LLM is typically trained on. Our hypothesis was that, even if these commercial models were available online in a location we did not know about, they were unlikely to be translatable directly to Python. 

\subsection{Case 1: critical care unit model}

The first case study was published in the Journal of Simulation \citep{ccu_case_study}. The model was coded in VBA and has never been published online to our knowledge.  The description of the model was published before any reporting guidelines for DES, but the paper contains a detailed description of the model and its parameters, although in some areas they are not reported in a manner that allows full replication of quantitative results reported in the paper (for example, an empirical distribution was used, but not detailed); although some obvious simplifications were available based on descriptions in the paper (e.g., use of a statistical distribution for elective inter-arrival time instead of an unreported empirical distribution).

\begin{figure}
    \centering
    \includegraphics[width=0.95\linewidth]{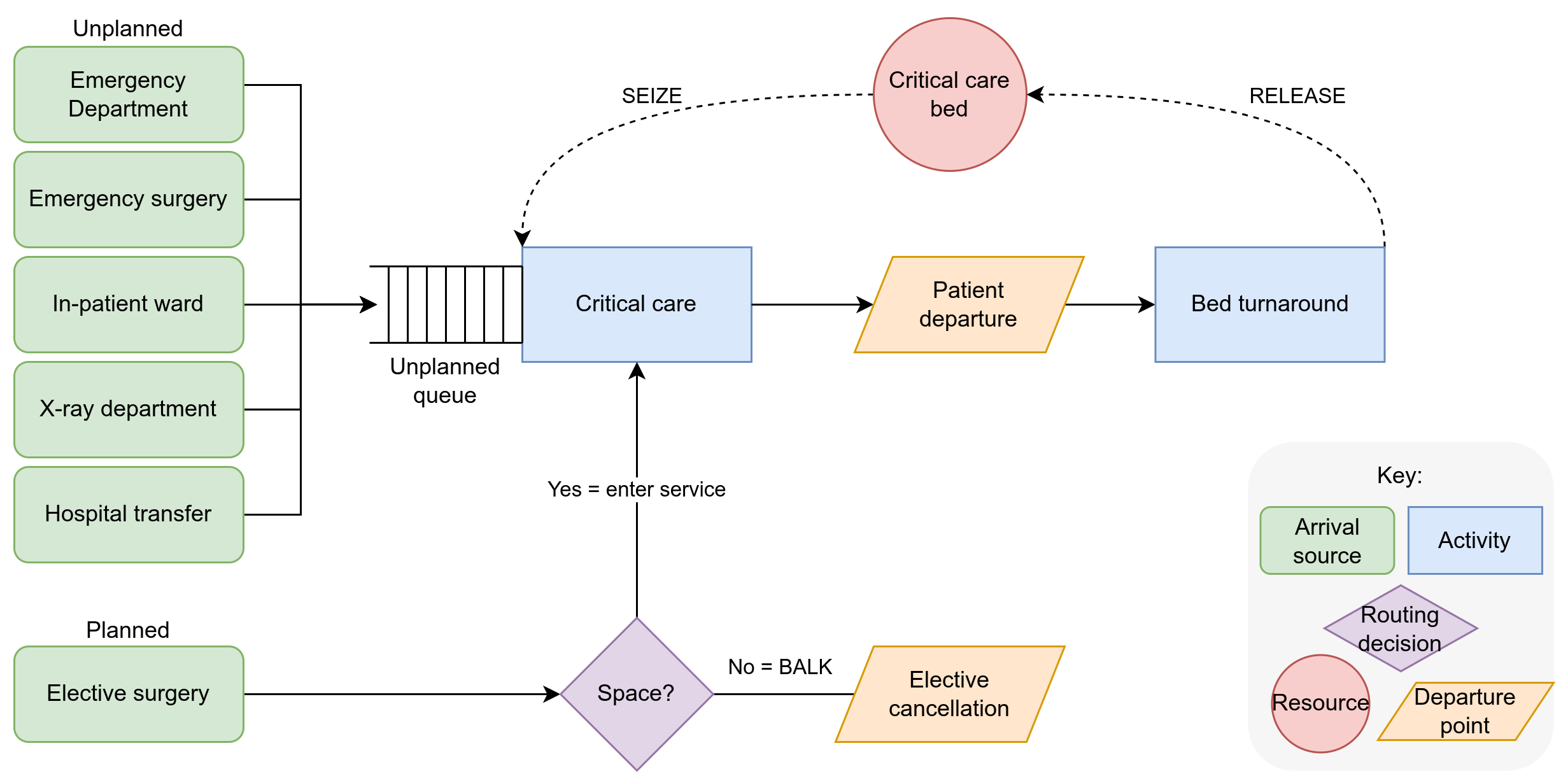}
    \caption{Logic diagram for the design of the CCU model}
    \label{fig:ccu_fig}
\end{figure}

The model represents a Critical (Intensive) Care Unit (CCU); we provide our interpretation of the process described in the article in Figure \ref{fig:ccu_fig}. It consists of six classes of entities that arrive following varying static distributions.  These arrivals are either unplanned (emergency) or planned (elective surgery), and share a total of 24 beds.  Unplanned emergency patients are prioritised for critical care beds.  Elective patient balk (a cancelled elective operations) if no beds are available. Patient classes have their own treatment time distributions (length of stay in the CCU).  After discharge a deterministic bed turnaround time is included to allow for intensive cleaning. The study aimed to explore capacity requirements and related scenarios and their impact on the number of cancelled operations. A warm-up period and multiple replications are employed.

\subsection{Case 2: stroke pathway capacity planning model}

The second case study was published in BMC Health Services Research \citep{stroke_case_study}. The model was coded in the Simul8 simulation package and has never been published online. The authors present a simple generic model to support health services to plan the capacity of an acute stroke ward, rehabilitation ward, and (optionally), early support discharge capacity (ESD). The model was published before reporting guidelines for DES, but contains a detailed appendix allowing for recreation of the model (and uses simple parameters and equations). 

\begin{figure}
    \centering
    \includegraphics[width=0.95\linewidth]{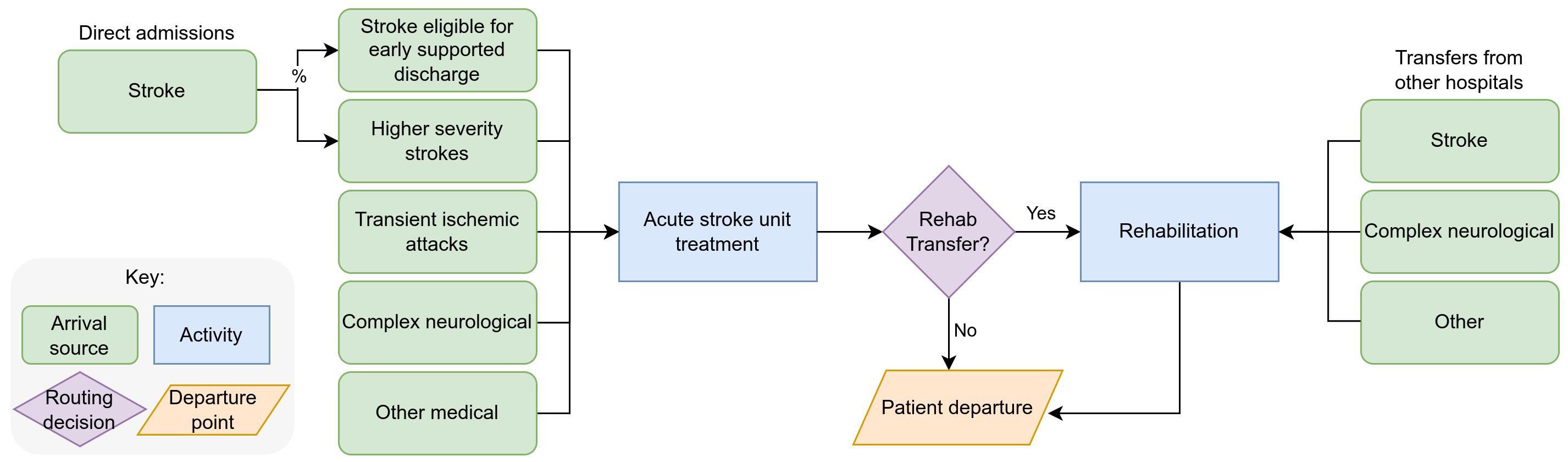}
    \caption{Logic diagram for design of the acute stroke and community rehabilitation capacity planning model}
    \label{fig:stroke_fig}
\end{figure}

The model allows users to specify a population of stroke, transient ischaemic attack (TIA), complex neurological, and other patient types who use acute and rehabilitation services. We provide our interpretation of the process described in the article in Figure \ref{fig:stroke_fig}. The patient classes have their own external inter-arrival distributions to acute and rehabilitation services, transfer probabilities between services, and length of stay distributions (where first sub-division occurs to model ESD versus non-ESD patients). The model takes an infinite capacity approach to capacity planning and estimates the probability of delay. A warm-up period and multiple replications are employed.  It has a clear logic diagram and documentation of parameters in the main article and an online appendix.  The ESD modelling is not used or documented in the article; we therefore chose to simplify the design and focus on the acute stroke and rehabilitation units, although we aimed for model setup to be simple to extend for ESD capacity modelling.

\section{Model design}

\subsection{\textit{SimPy} models}

All generated models were built in \textit{SimPy} \citep{simpy}: a process-based DES package implemented in Python that is available under a permissive MIT license. DES models are built by defining Python generator functions and logic to request and return resources. The package \textit{SimPy} is lightweight and provides a full event scheduling engine. Statistical distributions for sampling, common random number streams, output analysis tools, user interfaces, and model animation are not included. The implementation of all of these tools are available in the general and scientific Python stack. For example, \textit{NumPy} and \textit{pandas} for sampling and output analysis, and \textit{Matplotlib} for visualising results. For those unfamiliar with \textit{SimPy} we provide some sample code in Appendix \ref{app:simpy_example} Listing \ref{code:example_simpy}.

\subsection{Model coding plan}

The number of modelling iterations needed to recreate the models was not known in advance. Instead, we read the publications reporting the model designs and constructed a general plan that ordered and batched iterations into 12 aims of model recreation (e.g. each aim might take 2-3 prompts and code iterations to achieve). Table \ref{tab:model-building} details the ordered aims of the model recreation process along with a description and examples of changes to the model that could be expected.  Our aims took us from modelling of arrivals of patients and patient classes (e.g., types of stroke, or unplanned emergencies versus elective patients) through to a user interface allowing for basic experimentation. We believe this mirrors how the recreation of a DES model would take place regardless of whether an LLM was used. 

\begin{table}
\footnotesize
\centering
\caption{Ordered aims of the model recreation process}
\begin{tabular}{|l|p{4cm}|p{7cm}|}
\hline
\textbf{Aim} & \textbf{Description}                          & \textbf{Example additions to model}                                                                           \\ \hline
1            & Arrival processes and logic                   & Single and multiple classes of patients                                                                       \\ \hline
2            & Initial queuing logic and activities (delays) & Reneging, patient class dependent length of stay in a ward                                                    \\ \hline
3            & Separation of parameters from model logic     & A configurable Experiment class to hold all model parameters.         \\ \hline
4            & Simulated trace control                       & Functionality to hide and display simulated events                                                            \\ \hline
4            & Patient routing                               & Sampling to determine post-ward destinations and activities                                                   \\ \hline
5            & Additional queuing and activities             & Additional treatment in a different hospital/ward                                                             \\ \hline
6            & Results collection                            & Audit and calculate ward occupancy, bed utilisation, waiting times etc.                                       \\ \hline
7            & Warm-up period                                & Split the model run length into warm-up and results collection, reset all KPIs, introduce auditing processes. \\ \hline
8            & Multiple replications                         & Multiple unique runs of the model.                                         \\ \hline
9            & Output analysis procedures                    & Charts and summary tables                                                                                     \\ \hline
10           & Common random numbers                         & Allocate unique random number stream to each distribution                                                     \\ \hline
11           & User interface                                & A web browser-based interface for the model.                              \\ \hline
12           & Final bug fixes                               & Patch any remaining bugs identified by a 2nd modeller                                                         \\ \hline
\end{tabular}
\label{tab:model-building}
\end{table}

To optimise the organisation and usability of the \textit{SimPy} simulation model we adopted the approach of \cite{monks2023improving} in aims 3, 8 and 11.  The result is that model logic is separated from parameters using an \verb|Experiment| class (used to set up ``what-if" experiments).  The \verb|Experiment| class is used in combination with a multiple replications wrapper function to generate results.  This simple organisation enables quick integration with Python web app frameworks such as \textit{Streamlit} to make models usable by a wider group of people.

To enable both repeatable replications and variance reduction between experiments, we chose to implement common random number (CRN) streams in our models; i.e. each random statistical distribution used for sampling has its own unique controllable pseudorandom number stream \citep{Davies2007}.  This is in line with Case Study 2 which used Simul8 and implemented CRN. However, we note that Case Study 1 was implemented in VBA and it is unclear if CRN streams were implemented by the authors. We aimed to manage all random sampling through the \textit{numpy.random} module and the PCG-64 pseudorandom number generator \citep{numpy_random_generator}. We followed a simple approach where the replication number was used to spawn $n$ independent random number streams.

Python code should follow coding standards such as PEP8 and PEP257.  We chose to relax these standards to reduce the number of lines the LLMs had to generate (in terms of line wrapping and documentation). After all iterations were completed we used the tool \textit{black} to autoformat the code to meet PEP8 standards.

\section{Prompt Engineering}

\subsection{Prompts versus academic article write-up}

A naïve approach to recreating DES models with LLMs is to directly copy-and-paste text from the manuscript (along with instructions to translate the logic into a Python model) into a chat interface.  Although potentially time efficient, we argue that this is not an effective or reliable approach due to both the implicit healthcare process and modelling knowledge that human written narrative may contain. As a simple example, consider Box \ref{box:prompt-engineering} where the first paragraph is an excerpt from section 2.1 of the CCU case study \citep{ccu_case_study} that describes balking of elective patients in a CCU.

A human reader of this text may understand the context and logic of this text exactly; especially if they have viewed a logic diagram of the model beforehand. On the other hand, an LLM prompted using this text is less likely to produce consistent models due to the mixing of patient types in the discussion and lack of precision in the language.

The second half of Box \ref{box:prompt-engineering} constructs (engineers) a prompt for the LLM with the aim of being more precise. Instead of a discursive prose, we re-frame the natural language to:

\begin{enumerate}
    \item Instruct the LLM to add a new arrival source to the model for elective surgery patients;
    \item Precisely define which resources are used by the elective patients on arrival and if these are shared with other types of patients;
    \item Identify the outcome logic if all beds are in use versus if they are not all in use.
\end{enumerate}

\begin{textbox}[h!]
\begin{textexcerpt}

\textbf{Excerpt from journal article:}  \textit{If an arriving patient finds that all beds are occupied, they are sent to a queue. There are two queues built into the model, the `Unplanned Admissions' queue and the `Planned Admissions' queue. The patients in the `Planned Admissions' queue—that is the Elective surgery patients—have their surgery cancelled and are then sent home. The patients in the `Unplanned Admissions` queue wait until a bed becomes available.}  \\

\textbf{Engineered Prompt:} \\

Add a new arrival source to the CCU: Elective surgery patients. Elective surgery patients are modelled as a separate process from the unplanned admissions but share the critical care bed resources. \\

As an elective patient arrives at the CCU, a check is made on the number of critical care beds available.  There are two outcomes from this check:

\begin{itemize}
    \item  \textbf{Outcome 1}: The number of beds in use is equal to the total number of beds available. In this case, the elective patient leaves the model immediately. This is called a ``cancelled operation" event and should be reported to the user.

    \item \textbf{Outcome 2}. The number of beds in use is less than the total number of beds available. In this case, the elective patient requests a critical care bed, is treated, and is then discharged. 
\end{itemize}

\end{textexcerpt}
\caption{Example of prompt engineering. Excerpt taken from section 2.1 \citep{ccu_case_study}.}
\label{box:prompt-engineering}
\end{textbox}

\subsection{Common tokens}

In a healthcare simulation study, stakeholders and modellers may use multiple terms to refer to the same concept.  For example, the terms ``treatment time" and ``length of stay" may be used interchangeably in a conversation or a written article.  We aimed to make our prompts as specific as possible to obtain the iteration of the model that met our design. We therefore attempted to use a common token throughout an individual prompt and across iterations. We did allow for the shortening of tokens within prompts. For example, if we had introduced the concept of ``critical care bed" resources and the model had no other ambiguous resource names, we allowed our prompts to refer to ``beds".

\subsection{Initial prompts}

In both cases our initial prompts were designed to generate a simple working simulation model in \textit{SimPy} that generated patient arrivals only. Our hypothesis was that the level of detail and scope could then be expanded in further iterations. Our initial prompt was one of the most detailed provided. We broke it down into four sections. 

\begin{itemize}
    \item \textbf{Main command}: Specified the context (e.g., a critical care unit DES model), programming language, simulation package and that this was a code generating task.
    \item \textbf{General model logic}: Including time units, arrival sources, model boundaries and run length.
    \item \textbf{Simulation inputs}: For the first iteration this was always inter-arrival distributions and parameters.
    \item \textbf{Simulation methodology}: Underlying sampling tools and how this should be implemented.
\end{itemize}

To illustrate this method we include Box \ref{box:prompt_one}: the initial prompt from the CCU case study.

\begin{textbox}[h!]
\begin{textexcerpt}

\textbf{Main command:} Code a discrete-event simulation model of a critical care unit (CCU) in Python 3.10 and \textit{SimPy} 4.  Code the full model specified. Do not return a simplified version. Show all code. \\

\noindent\textbf{General model logic:}

\begin{itemize}
    \item All time units in the model are in hours.
    \item Each patient in the model has a unique identifier.  The first patient to arrive has an identifier of 0. For each subsequent patient increment the identifier by 1.
    \item Patients to arrive at the CCU from five different sources: Accident and Emergency, the Wards, Emergency surgery, other hospitals, or the X-Ray department. 
    \item Each source has a different inter-arrival time distribution. 
    \item After patients arrive they immediately leave the model.
    \item All patient types must have their own generator function
    \item The model should print out useful information after each event.
    \item The model should include a user-settable run length. This defaults to 12 months.
\end{itemize}

\noindent\textbf{Simulation inputs:} The inter-arrival time distributions and parameters of patients are dependent on patient type. For each distribution, time is measured in hours.

\begin{itemize}
    \item Accident and Emergency = Exponential: 22.72
    \item The Wards = Exponential: 26.0
    \item Emergency surgery = Exponential: 37.0
    \item Other hospitals = Exponential: 47.2
    \item The X-Ray department = Exponential: 575.0
\end{itemize}

\noindent\textbf{Simulation methodology:} NumPy should be used for sampling. Each inter-arrival distribution should have its own numpy.random.Generator object.

\end{textexcerpt}
\caption{Initial prompt example: CCU case study.}
\label{box:prompt_one}
\end{textbox}

\subsection{Refactoring prompts}

Within an iteration, we occasionally made use of subsequent short prompts to refactor the code closer to our requirements or expectations. For example, if the code generated a set of functions when we preferred classes we would issue the follow-up prompt: ``re-factor the functions into a CCU class".

\subsection{One-shot prompt engineering}

When refactoring of code was judged to be complex we chose to provide a short example of the refactoring in the prompt for the LLM to mimic (so-called ``1-shot prompt engineering").  

An excerpt from a prompt given to the LLM from the stroke case study is provided in Box \ref{box:prompt-one-shot} where we are refactoring the code to use random sampling that follows best practice. The LLM must modify the code in multiple methods and also count from 0 to the number of random streams implemented. We therefore provide an example of how this should be implemented.

\begin{textbox}[h!]
\begin{textexcerpt}

Modify the \verb|acute_treatment| functions in \verb|AcuteStrokeUnit| class. Do not modify the $\verb|acute_treatment|$ functions.

Code that uses \verb|numpy.random|, must be replaced with a call to a unique stream in the \verb|Experiment| list \verb|streams|. Select the stream using a hard-coded integer. \\

Start from zero and increment by 1 each time to allocate a unique number to each stream.  \textbf{E.g.} In \verb|stroke_acute_treatment| the first instance of \verb|length_of_stay = np.random.lognormal(mu, sigma)| becomes\\
\verb|length_of_stay = self.experiment.streams[0](mu, sigma)|; the second instance uses index 1 and the third uses index 2 and so on.

\end{textexcerpt}
\caption{Example of one-shot prompt engineering to guide code refactoring.}
\label{box:prompt-one-shot}
\end{textbox}

\subsection{Numbered steps}

Many simple functions in programming are a series of steps to be followed by the Python interpreter. Where they were very clear, we specified them as a number-ordered list of natural language instructions that the LLM could follow. Box \ref{box:numbered_steps} illustrates such a prompt applied to create an initial iteration of a Streamlit interface for the stroke case study model.

\begin{textbox}[h!]
\begin{textexcerpt}

Write python code that creates an interactive user interface using the package Streamlit. The interface should include a main window. The main window contains a button labelled ``Simulate”. After the button is pressed the following logic is implemented:

\begin{enumerate}
    \item Display a spinner with the text “please wait for results”.
    \item Run the python code included below \textit{[not shown]}.
    \item Display a Streamlit table for \verb|df_acute| and \verb|df_rehab| results.
    \item Display all plots.  Plotting functions return a tuple of figure, axis.
\end{enumerate}

All classes and functions should be imported from a module called \verb|stroke_rehab_model|.
\end{textexcerpt}
\caption{Example of specify a prompt using numbered steps.}
\label{box:numbered_steps}
\end{textbox}

\subsection{Restrictive clauses in prompts}

We appended restrictive clauses to our prompts to avoid changes to parts of the code that were not part of our design. For example, when specifying a modification to the treatment of patients we could append ``Do not modify the \verb|patient_generator| functions at all" to ensure these were not modified in the same way. Similarly, if we were interested in refactoring the \verb|Experiment| class to add new variables we might specify ``only modify the Experiment class" or ``do not modify the CCU class" to avoid small changes to the design between iterations.

\section{Testing of generated code}

Following each iteration of model generation a four-step testing procedure was employed.  

\subsection{Step 1: visual inspection of the code.}

Our initial approach was to use the JupyterLab Integrated Development Environment (IDE) to visually inspect the generated model and check for obvious logical bugs, unused code or package imports, outdated Python libraries, fabricated functionality (e.g., functions that do not exist), etc.  However, during Stage 1, we found that visual inspection became too difficult when a modification of existing code took place.

In iteration 11 of Case 2, we enhanced the code inspection process by including the use of a Python library called \textit{nbdime}\footnote{https://nbdime.readthedocs.io/en/latest/} that provided a highlighted difference between two versions of the same notebook. An example of a difference is illustrated in Listing \ref{code:diff} . Line 2 is labelled `-' and line 3 is labelled `+'. This shows the user line 2 has been edited: a new parameter \verb|post_asu_probabilities| has been added to the code. Line 8 is labelled `+' and shows a new line of code added to the model.  The use of \textit{nbdime} meant that we did not miss any modification that unexpectedly removed code or unexpectedly modified existing code from prior iterations. Our process was therefore updated to:

\begin{enumerate}
    \item Copy the prior iteration of the notebook;
    \item Replace any existing functions, classes or scripts with new versions generated in the iteration;
    \item Add in cells to hold new functions, classes, or scripts;
    \item Generate highlighted differences between the new notebook and the prior iteration.
\end{enumerate}

\begin{tcolorbox}[colback=gray!5, colframe=black, title=Model and code testing]

\begin{lstlisting}[language=python, caption=Example of differencing two iterations of the generated model, label=code:diff]
 class PatientType:
-    def __init__(self, name, interarrival_time):
+    def __init__(self, name, interarrival_time, post_asu_probabilities):
         self.name = name
         self.interarrival_time = interarrival_time
         self.count = 0
         self.rng = np.random.default_rng()
+        self.post_asu_probabilities = post_asu_probabilities
 
     def generate_interarrival_time(self):
         return self.rng.exponential(self.interarrival_time)
\end{lstlisting}

\end{tcolorbox}

\subsection{Step 2: classical verification of the simulation model}

We designed and conducted a series of experiments with the model.  These included:

\begin{enumerate}
    \item Extreme value tests (e.g limiting arrival types, zero or extreme lengths of stay).
    \item Varying parameters such as run length or the warm-up period.
    \item  Data collection and post-run processing.
    \item Basic unit test that any equations produce expected values.
    \item Testing of individual components within the model (for example, in Case 2 testing the Acute Stroke Unit separate from the Rehabilitation Ward).
    \item Visual inspection of plots produced.
    \item Inspection of summary statistics of performance measures.
    
\end{enumerate}

For each iteration of the model we re-ran prior model testing and \textit{if required} added new tests.

\subsection{Step 3: Creation of automated and manual tests}

After the completion of a model we stored all generated code in a dedicated Python module. We then refactored tests created in Step 2 into two sets of tests that were easy to run and included with the full models.  These were split into automated and manual tests.  The automated tests worked with the Python package \textit{pytest}. These tests have simple quantitative pass/fail criteria. The \textit{pytest} software automatically detects and runs all automated tests and reports successes and failures to a user.  The manual tests involve visual inspection (e.g., inspecting the simulated trace and charts).  

\subsection{Step 4: Testing by a second modeller}

The final step in testing was conducted by the second modeller. The modeller was provided with:

\begin{enumerate}
    \item The journal articles describing the two simulation models.
    \item The Python environment containing all software used to run the models.
    \item The Python module(s) containing all model code for the two case studies.
    \item The set of automated and manual tests.
    \item Jupyter notebooks that contained a (human-created) Python script for running the models/user interfaces and detailed usage instructions.
    \item All prompts (in sequence) used to generate code and all Jupyter notebooks containing the iterations of the model/testing.
\end{enumerate}
   
The second modeller reviewed all of this information and attempted to run the models and tests. We used this step to identify:

\begin{itemize}
    \item Missed errors in the model recreation (either by the LLM or the user).
    \item Typos, mistakes or copy-and-paste errors in iteration notebooks.
    \item Additional formal testing that should be conducted.
    \item Improvements that were needed in model documentation.
\end{itemize}

Any feedback from this review was addressed before proceeding to Stage 2 in our experimental procedure.

\section{Results}

\begin{figure}
    \centering
    \includegraphics[width=1\linewidth]{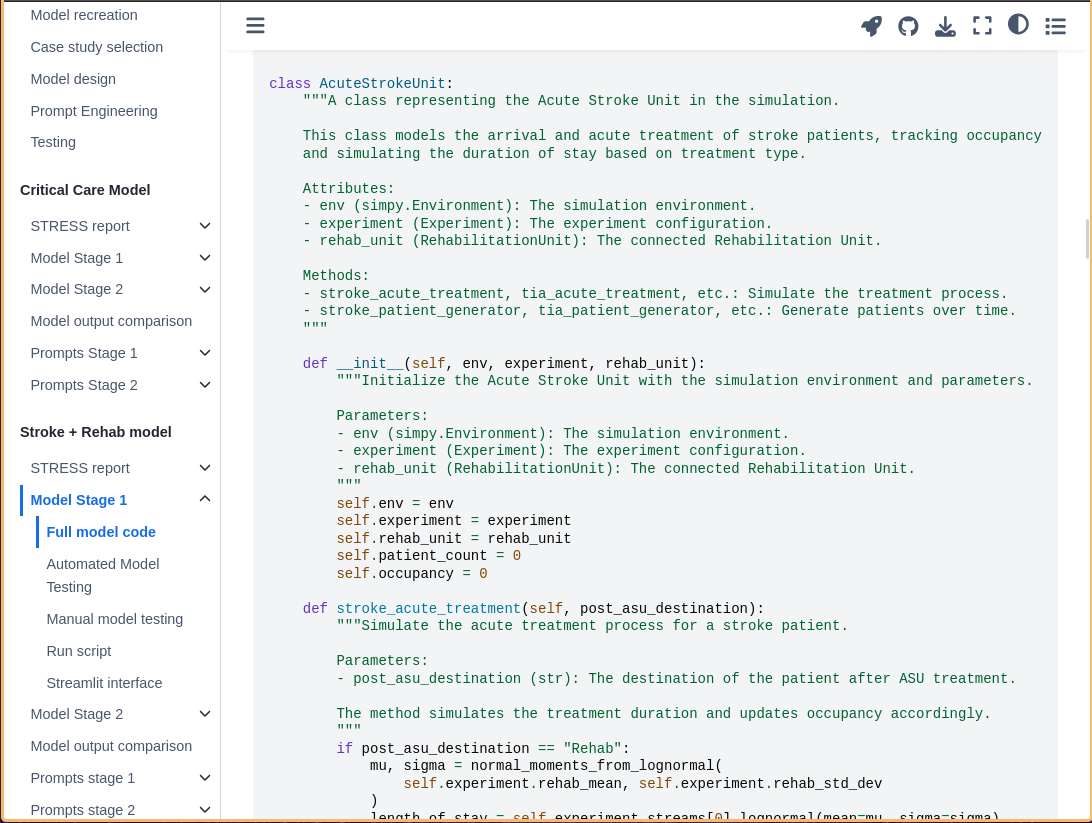}
    \caption{Screenshot of the study's research compendium. Model code, testing, and prompt archive are organised by case study and stage.}
    \label{fig:research_compendium}
\end{figure}

Table \ref{tab:artefacts} lists links to all of the research artefacts generated in the study. To preserve the outputs we have archived each artefact at Zenodo and provide a DOI.  Figure \ref{fig:research_compendium} is a screenshot of the study's online research compendium.  The compendium organises the generated model code, testing, and prompts by the case study and stage.  We have deployed usable versions of the generated models as a web app to GitHub Pages.

For each model we report the outputs generated by the model; the results of the internal replication test; describe the code generated, and provide a screenshot of the user interface.  The full code listings are available in the research compendium; here we provide short excerpts of the code in Listings.

\begin{table}[]
\small
\centering
\caption{Links to research artefacts and generated models.}
\begin{tabular}{|l|l|}
\hline
\multicolumn{2}{|c|}{\textbf{Research compendium}} \\ \hline
\textbf{Deployment} & \url{https://pythonhealthdatascience.github.io/llm_simpy} \\ \hline
\textbf{DOI} & \url{https://doi.org/10.5281/zenodo.15090962} \\ \hline
\textbf{GitHub} & \url{https://github.com/pythonhealthdatascience/llm_simpy} \\ \hline \hline
\multicolumn{2}{|c|}{\textbf{Generated models as a web app}} \\ \hline
\textbf{Deployment} & \url{https://pythonhealthdatascience.github.io/llm_simpy_models} \\ \hline
\textbf{DOI} & \url{https://doi.org/10.5281/zenodo.15082494} \\ \hline
\textbf{GitHub} & \url{https://github.com/pythonhealthdatascience/llm_simpy_models} \\ \hline
\end{tabular}
\label{tab:artefacts}
\end{table}

\subsection{Case 1: critical care unit model}

\subsubsection{Model outputs}

The simulation models generated in Stages 1 and 2 produced identical results to one another (to two decimal places), as in Table \ref{tab:ccu_comparison_table}. The table summarises the number of patient arrivals to the models, the four output measures in the design and how these vary across six experiments with the model.

The model results did not replicate those reported in \cite{ccu_case_study}. Our results have a higher arrival rate overall and higher occupancy of the CCU. The explanation would appear to be that we did not have access to information about the empirical distributions used for elective patients in the original article and the alternative distribution recommended in the article is not representative.

\begin{table}
\footnotesize 
\caption{Comparison of critical care model outputs: Stage 1 versus Stage 2 (internal replication). Figures are mean (standard deviation).}
\begin{tabular}{|p{1.2cm}|>{\raggedleft}p{3.5cm}|p{1cm}|p{1cm}|p{1cm}|p{1cm}|p{1cm}|p{1cm}|}
\hline
\textbf{Study Stage}              & \textbf{Metric}                                           & \textbf{23 beds} & \textbf{24 beds} & \textbf{25 beds} & \textbf{26 beds} & \textbf{27 beds} & \textbf{28 beds} \\ \hline
\multirow{5}{*}{\textbf{Stage 1}} & \textbf{0. Patient Count}                                 & 1,650.4 (17.83)  & 1,650.4 (17.83)  & 1,650.4 (17.83)  & 1,650.4 (17.83)  & 1,650.4 (17.83)  & 1,650.4 (17.83)  \\ \cline{2-8} 
                                  & \textbf{1. Cancelled Elective Operations}                 & 390.6 (30.57)    & 337.8 (38.75)    & 279.0 (39.13)    & 231.4 (33.83)    & 178.4 (32.46)    & 139.8 (27.58)    \\ \cline{2-8} 
                                  & \textbf{2. Bed Utilization}                               & 0.9 (0.02)       & 0.9 (0.02)       & 0.9 (0.02)       & 0.9 (0.02)       & 0.8 (0.02)       & 0.8 (0.02)       \\ \cline{2-8} 
                                  & \textbf{3. Bed Occupancy}                                 & 21.3 (0.49)      & 21.8 (0.50)      & 22.3 (0.54)      & 22.6 (0.56)      & 23.0 (0.58)      & 23.3 (0.62)      \\ \cline{2-8} 
                                  & \textbf{4. Mean Unplanned Admission Waiting Time (hours)} & 103.8 (72.08)    & 62.5 (55.23)     & 35.0 (29.28)     & 20.8 (15.59)     & 12.0 (7.66)      & 7.0 (3.76)       \\ \hline \hline
\multirow{5}{*}{\textbf{Stage 2}} & \textbf{0. Patient Count}                                 & 1,650.4 (17.83)  & 1,650.4 (17.83)  & 1,650.4 (17.83)  & 1,650.4 (17.83)  & 1,650.4 (17.83)  & 1,650.4 (17.83)  \\ \cline{2-8} 
                                  & \textbf{1. Cancelled Elective Operations}                 & 390.6 (30.57)    & 337.8 (38.75)    & 279.0 (39.13)    & 231.4 (33.83)    & 178.4 (32.46)    & 139.8 (27.58)    \\ \cline{2-8} 
                                  & \textbf{2. Bed Utilization}                               & 0.9 (0.02)       & 0.9 (0.02)       & 0.9 (0.02)       & 0.9 (0.02)       & 0.8 (0.02)       & 0.8 (0.02)       \\ \cline{2-8} 
                                  & \textbf{3. Bed Occupancy}                                 & 21.3 (0.49)      & 21.8 (0.50)      & 22.3 (0.54)      & 22.6 (0.56)      & 23.0 (0.58)      & 23.3 (0.62)      \\ \cline{2-8} 
                                  & \textbf{4. Mean Unplanned Admission Waiting Time (hours)} & 103.8 (72.08)    & 62.5 (55.23)     & 35.0 (29.28)     & 20.8 (15.59)     & 12.0 (7.66)      & 7.0 (3.76)       \\ \hline
\end{tabular}
\label{tab:ccu_comparison_table}
\end{table}

\subsubsection{Model code}
Disregarding comments and documentation, Stage 1 generated a model consisting of 262 lines of code and Stage 2 generated 355 lines of code. Both models passed the same batch of 28 verification tests. The difference in the design of the Python classes representing an experiment and CCU model logic can be seen in the number of class attributes and methods in Table \ref{tab:ccu_code}.

The final code files from Stages 1 and Stage 2 (our internal replication) for the critical care unit model and its interface were overall very similar. For example, code listings \ref{code:ccu1} and \ref{code:ccu2} illustrate the similarities in code generated to model emergency department arrivals. Minor differences included the naming of variables, functions and classes. Another minor difference was the setup of random number generators for each activity in the model.  However, both approaches in Stage 1 and 2 were acceptable.  

A more substantial difference is that the Stage 2 code is arguably easier to understand for a new user than Stage 1 code.  For example, the LLM in Stage 2 generated an \verb|Experiment| class where each parameter used in a statistical distribution (e.g. the mean inter-arrival times) was implemented as a named variable. Whereas in Stage 1 code the LLM generated an \verb|Experiment| class where inter-arrival means were set via a list of unnamed parameter values. This increased clarity resulted in more lines of code in Stage 2 than in Stage 1; although we do not consider this a good or bad outcome.

A similar difference in clarity can be seen in the code to convert the mean and standard deviation of a log-normal into scale and shape parameters (suitable for the \textit{numpy} log-normal functions). In Stage 2 the logic was (optimally) implemented in a reusable function.  In Stage 1 the conversion logic is coded directly into each process and is harder to follow and test. 

\begin{tcolorbox}[colback=gray!5, colframe=black, title=Example code generated in Stage 1 and Stage 2 (Critical Care Model)]

\begin{lstlisting}[language=python, caption=Example CCU code from Stage 1, label=code:ccu1]
class CCUModel:
    def __init__(self, env, experiment):
        self.env = env
        self.experiment = experiment
        self.patient_count = 0
        self.critical_care_beds = simpy.Resource(
            env, capacity=experiment.num_critical_care_beds
        )

    def patient_arrival_AE(self):
        while True:
            yield self.env.timeout(
                self.experiment.streams[0].exponential(
                    self.experiment.interarrival_means[0]
                )
            )
            self.patient_count += 1
            if self.experiment.trace:
                print(
                    f"Patient {self.patient_count} ... time {self.env.now}"
                )
            self.env.process(
                self.unplanned_admission(
                    self.experiment.stay_distributions[0]
                )
            )
\end{lstlisting}

\vspace{1em} 

\begin{lstlisting}[language=python, caption=Example CCU code from Stage 2, label=code:ccu2]
class CCU:
    def __init__(self, env, experiment):
        self.env = env
        self.experiment = experiment
        self.patient_id_counter = 0
        # ...
        self.critical_care_beds = simpy.Resource(
            env, capacity=self.experiment.num_critical_care_beds
        )

    def accident_emergency_arrivals(self):
        while True:
            yield self.env.timeout(
                self.experiment.rng_accident_emergency.exponential(
                    self.experiment.accident_emergency_arrival_rate
                )
            )
            self.patient_id_counter += 1
            if self.experiment.trace:
                print(
                    f"Patient {self.patient_id_counter} ...{self.env.now:.2f}"
                )
            self.env.process(
                self.unplanned_admissions_process(
                    "Accident and Emergency"
                )
            )
\end{lstlisting}

\end{tcolorbox}

\begin{table}[htbp]
\footnotesize
\centering
\caption{Count of CCU model code components comparing Stages 1 and 2.}
\label{tab:ccu_code}
\begin{tabular}{|l|p{2cm}|p{2cm}|p{2cm}|p{2cm}|}
\hline
\textbf{Component} & \multicolumn{2}{c|}{\textbf{Number of Attributes}} & \multicolumn{2}{c|}{\textbf{Number of Methods/Functions}} \\ \cline{2-5}
                    & \textbf{Stage 1} & \textbf{Stage 2} & \textbf{Stage 1} & \textbf{Stage 2} \\ \hline
\verb|Experiment| class    & 13               & 27               & 3                & 2                \\ \hline
\verb|CCU model logic| class & 4                & 9                & 10               & 12               \\ \hline
Functions            & N/A              & N/A              & 6                & 6                \\ \hline
\end{tabular}
\end{table}

\subsubsection{Prompts}

In total, 22 iterations of the model were used to build the model and interface. We report the number of prompts by iteration and stage in Appendix \ref{app:ccu_results} Table \ref{app:ccu_prompt_numbers}. In Stage 1 there were 26 prompts passed to the LLM. The number of prompts increased to 36 in Stage 2. Five of the 10 extra prompts occurred in the first two iterations of the model. Some minor additional prompting was needed to ensure comparable performance measures.  The final iteration of the model was a bug fix that was only relevant to Stage 1; therefore Stage 2 saved one prompt.

\subsection{Case 2: stroke pathway capacity planning model}

\subsubsection{Model outputs}

The results of the two generated simulation models were identical to 2 decimal places. The results for the Stage 1 and Stage 2 models are reported and compared graphically in Figure \ref{fig:stroke_comparison_acute} and Figure \ref{fig:stroke_comparison_rehab} (in Appendix \ref{app:ccu_prompt_numbers}). The figures show that the probability of delay and ward occupancy match across the acute and rehabilitation wards within the 2 models.

The outputs from the generated models results replicated the results reported in the original article \citep{stroke_case_study}; although we note that we did not run all of the experiments reported in the article.

\begin{figure}
    \centering
    \includegraphics[width=1\linewidth]{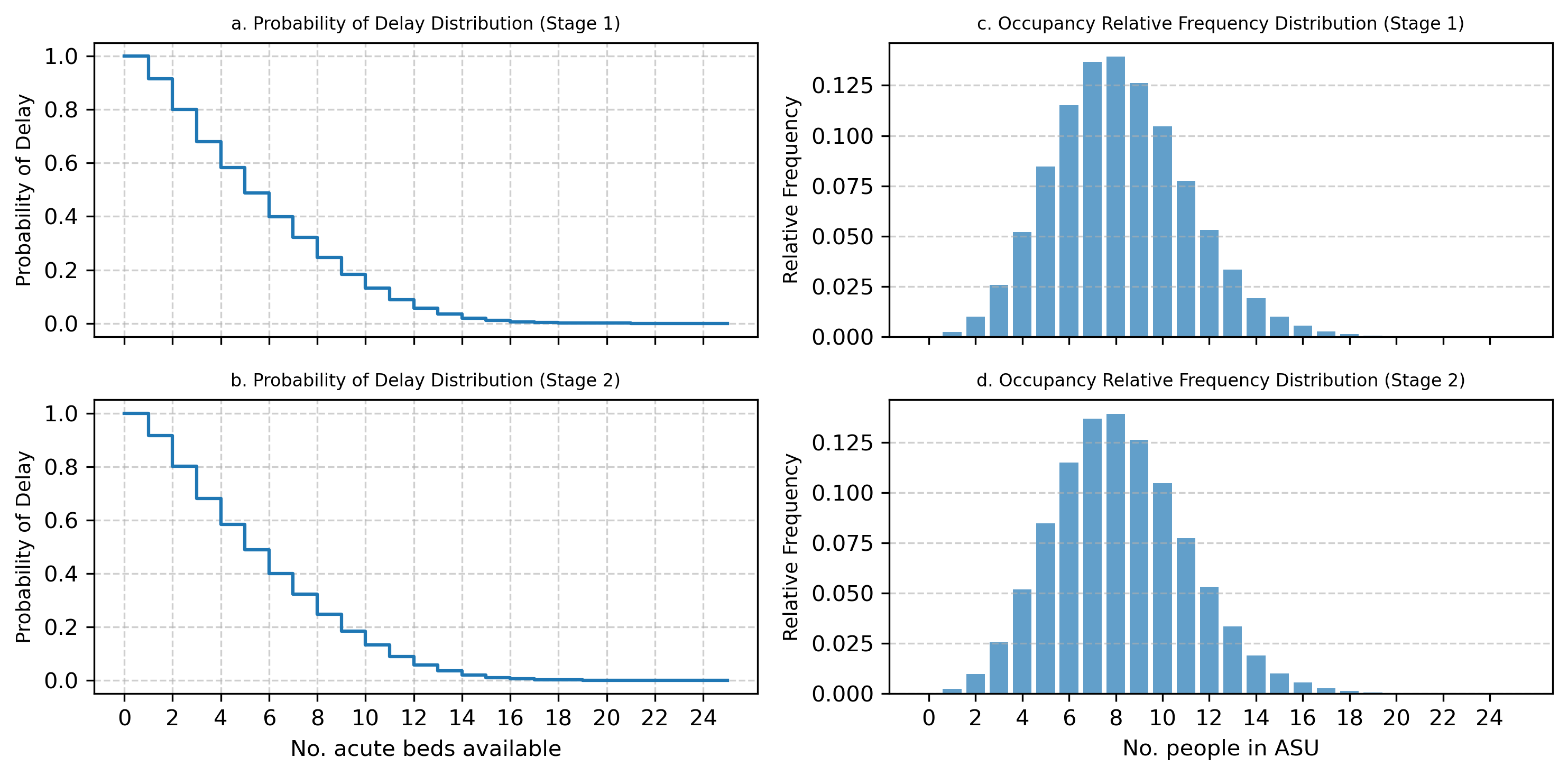}
    \caption{Acute stroke unit (ASU) outputs: comparison of Stage 1 and Stage 2 models.}
    \label{fig:stroke_comparison_acute}
\end{figure}

\subsubsection{Model code}

Disregarding comments, documentation and the Streamlit interface, Stage 1 generated a \textit{SimPy} model consisting of 436 lines of code and Stage 2 generated 531 lines of code.  Both models passed the same batch of 34 verification tests.

The final code files from Stage 1 and Stage 2 for the stroke capacity planning model have some substantial differences.  Table \ref{tab:stroke_code_comparison} summarises the difference in implementation. Notable differences included an additional \verb|PatientType| class in the Stage 2 model; and a six-fold difference in the number of member attributes of the \verb|Experiment| class.

\begin{table}[htbp]
\footnotesize
\centering
\caption{Count of stroke capacity planning model code components comparing Stages 1 and 2.}
\label{tab:stroke_code_comparison}
\begin{tabular}{|l|p{2cm}|p{2cm}|p{2cm}|p{2cm}|}
\hline
\textbf{Component} & \multicolumn{2}{c|}{\textbf{Number of Attributes}} & \multicolumn{2}{c|}{\textbf{Number of Methods/Functions}} \\ \cline{2-5}
                                       & \textbf{Stage 1} & \textbf{Stage 2} & \textbf{Stage 1} & \textbf{Stage 2} \\ \hline
\verb|Experiment| class                 & 36               & 5                & 3                & 5                \\ \hline
\verb|PatientType| class                & N/A              & 6                & N/A              & 3                \\ \hline
\verb|AcuteStrokeUnit| model logic class & 5                & 7                & 6                & 7                \\ \hline
\verb|RehabilitationUnit| model logic class & 6               & 18               & 7                & 9                \\ \hline
Functions (excluding classes)           & N/A              & N/A              & 11               & 10               \\ \hline
\end{tabular}
\end{table}

The difference in the code led to a difference in the how each model was setup to run an experiment. For example, in the stage 1 model the code to setup an experiment that simulated a 5\% increase in stroke patients, and then check the parameter value is shown in Listing \ref{code:stroke_s1}. The equivalent code in stage 2, shown in Listing \ref{code:stroke_s2} involved an additional line of code to create a Python dictionary and a collection data-structure to access the internal parameters.

We do not argue that either of the approaches generated by the LLM is optimal. Rather that there are pro’s and con’s to their implementations. Stage 1 code offers a simple code interface, but does not choose a clear naming convention (i.e. \verb|stroke_mean| is not specific to inter-arrival time). Stage 1 also does not clearly separate model parameters from the outputs of the experiment. Stage 2 code requires more code and requires a user to understand Python dictionaries. Stage 2’s hierarchy to access parameters is more complex than stage 1’s (including the internal workings of Experiment), but it uses clear specific naming conventions for patients types and their different parameters configurations.

\begin{tcolorbox}[colback=gray!5, colframe=black, title=Comparison of Python code to setup Experiments in Stages 1 and 2]

\begin{lstlisting}[language=python, caption=Setup a stroke model experiment in Stage 1, label=code:stroke_s1]
# setup experiment
default_experiment = Experiment(stroke_mean=1.2*1.05)

# access and check parameter value
print(default_experiment.stroke_mean)

\end{lstlisting}

\vspace{1em} 

\begin{lstlisting}[language=python, caption=Setup a stroke model experiment in Stage 2, label=code:stroke_s2]
# setup paramater dictionary
experiment_params = {
    "patient_types": {"Stroke": {"interarrival_time": 1.2 * 1.05}}
}

# pass to Experiment. LLM provided code that updates internal parameter dicts
demand_experiment = Experiment(experiment_params)

# access and check parameter value
print(
    demand_experiment.params['patient_types']['Stroke']['interarrival_time']
)
\end{lstlisting}

\end{tcolorbox}

\subsubsection{Prompts}

In total 31 iterations of the model were used to build the model and interface. In Stage 1 this consisted of 41 prompts passed to the LLM. The number of prompts increased to 57 in Stage 2. An additional prompt was needed in Stage 2 to fix a variable type bug introduced by the LLM for representing “patient type” across the acute and rehab sections of the model. Stage 2 required 4 additional prompts for introducing common random numbers streams to the LLM struggling to assign streams across model activities.  The number of prompts are report in Appendix \ref{app:ccu_results} Table \ref{app:stroke_prompt_numbers}.

\subsection{Model Interfaces}

In both case studies, we were successful in generating model interfaces using the streamlit Python package. Figure \ref{fig:combined_interfaces} depicts the model interfaces from stage 1 (we include stage 2 figures in the Appendix \ref{app:interfaces} Figure \ref{fig:combined_interfaces_stage2}). The generated interfaces were largely consistent across stage 1 and 2. This was possibly due to the simplicity of our requirements.  A minor difference was observed between stage 1 and 2 in the second case study.  In Figure \ref{fig:stroke_interface_stage1} Perplexity generated input widgets used to manipulate the model input parameters. In Stage 1 “numeric boxes” were used where-as in stage 2 “sliders” are used. The former allowed the input parameters to take negative values - that result in a model error (as you cannot have a negative inter-arrival arrival rate) - where-as sliders prevented this unacceptable setting with a minimum value (although still allowed 0).  This result suggests that prompts should specify valid ranges for input widgets.

\begin{figure}
    \centering
    \begin{subfigure}{\linewidth}
        \centering
        \includegraphics[width=0.75\linewidth]{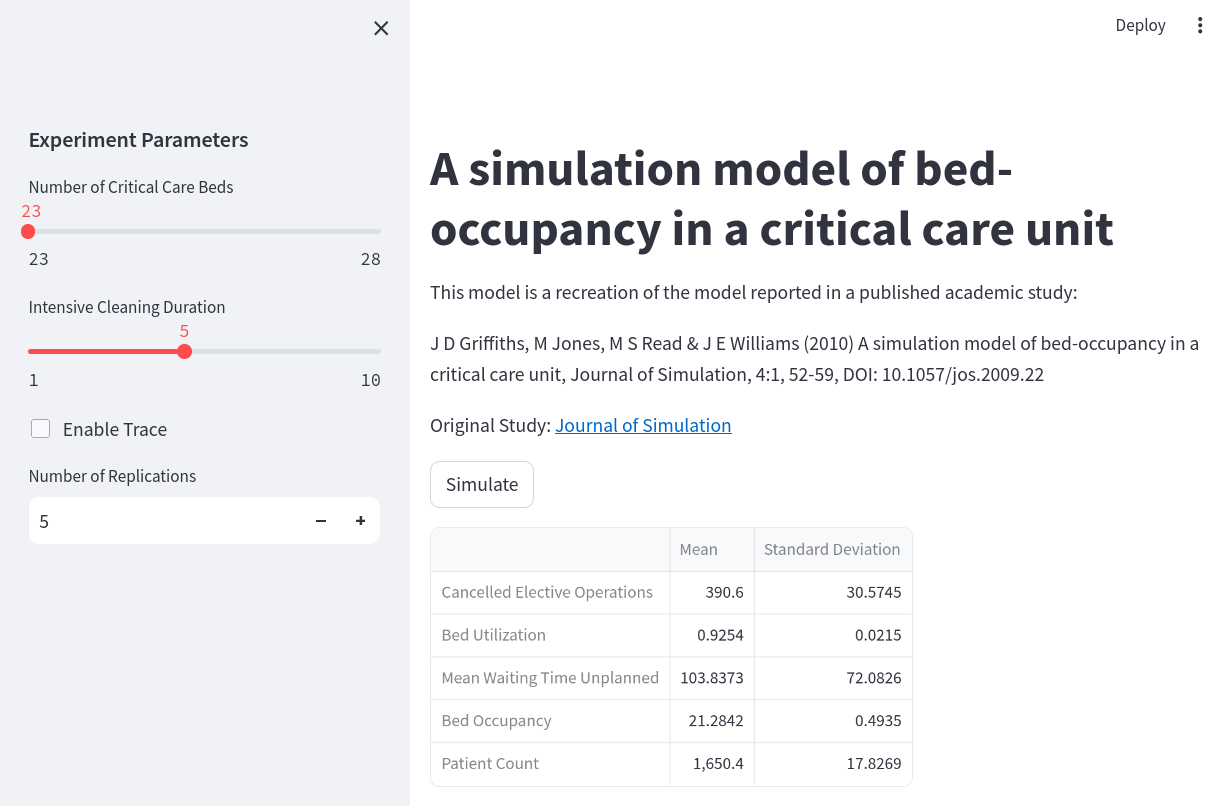}
        \caption{Generated streamlit interface for Critical Care Unit}
        \label{fig:ccu_interface_stage1}
    \end{subfigure}
    
    \vspace{1cm}  
    
    \begin{subfigure}{\linewidth}
        \centering
        \includegraphics[width=0.75\linewidth]{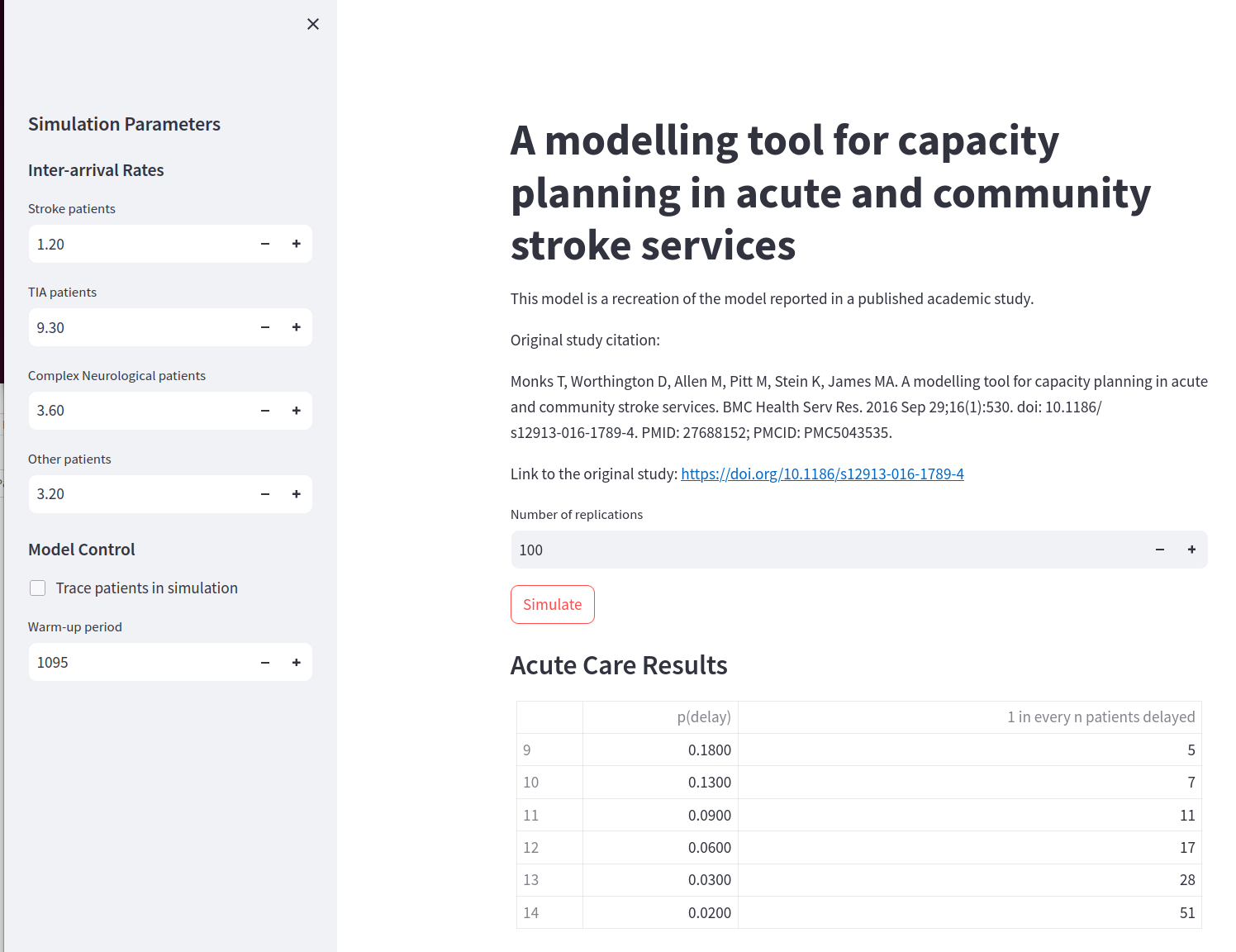}
        \caption{Generated interface for the stroke capacity planning model}
        \label{fig:stroke_interface_stage1}
    \end{subfigure}
    
    \caption{Model interfaces from stage 1}
    \label{fig:combined_interfaces}
\end{figure}
\section{Discussion}

Our objective was to test if we could recreate a version of published healthcare DES models in FOSS using generative AI. We selected two DES models from the literature where the coded model was not available and the original model was reported to have been implemented in commercial software that could not have formed part of the internet training data for a LLM. We read the natural language description of models in these papers and formed a model design. Based on model designs, we engineered prompts for the free tier of a generative AI tool (Perplexity) to code a Python and \textit{SimPy} model and a basic \textit{Streamlit} browser interface. We found that using our iterative approach to coding the model we were able to generate relatively sophisticated versions of two DES models. By sophisticated we mean:

\begin{itemize}
    \item The model logic was represented by up to 600 lines of Python code representing a queuing network in healthcare; this is relatively sophisticated compared to simulation models generated by others, for example, models were only represented by 20-30 lines of Python code \citep{jackson_2024,plooy_ai_2023,frydenlund2024modeler};
    \item The generated model design followed best practice recommendations \citep{monks2023improving} i.e. we separated model parameters from logic to enable simple experimentation, automated model testing, and linkage to a user interface;
    \item Between the two models we included multiple arrival sources and entity classes, balking, sampling mechanisms, queuing network routing, and calculation of multiple performance measures;
    \item The models' run length and output calculation were split into warm-up and data collection periods;
    \item Sampling in the models was implemented using streaming (common random numbers) and results across replications were repeatable.
\end{itemize}

We were able to reproduce our findings internally using the original prompts and several additional smaller prompts. In each case, the model code generated was different but produced consistent results. Given our ambition to preserve the time and effort put into coded models reported in the literature, we have archived all of our generated models in the Zenodo open science archive. This provides a guarantee on the persistence of the coded models - as long as the European Organization for Nuclear Research (CERN) exists. The models are MIT-licensed and available for others to download, inspect, reuse, adapt and redistribute with citation.

We would emphasise that, although successful in both stages of the study (the original recreation of a model design and its internal reproduction), neither were straightforward and provided several prompt engineering, model testing, and usage challenges for generative AI.  We argue that we were only successful due to the process we followed: our use of iterations to add model detail and scope, our approach to differencing and testing code, and our use of modellers experienced in DES, healthcare, and FOSS. We do not believe it would have been achievable without all of these elements. In the following section, we describe the challenges and opportunities of generative AI for recreating model designs.

\subsection{Challenges generating models}

\subsubsection{Lazy generation}

We quickly found that Perplexity used a strategy to reduce the amount of Python code it had to generate when prompted to modify code. For example, Python classes were partially generated with sections of code replaced by comments such as ``remaining functions go here" or ``original method code continues here". We can only speculate why the tool behaved this way given our prompts, but we might assume that less text generated required less computation and cost for the provider. We were able to by-pass this undesirable behaviour by a.) appending the clause ``Show all of the model code";  b.) prompting for specific(s) part of the model to be generated e.g., ``Show the full \verb|RehabilitationUnit| including all patient generator and treatment functions" (e.g., Stage 1 iterations 15 and 27 of the stroke model); c.) using a restrictive clause to only generate functions and classes that were modified e.g., ``only show the code for these three functions" (e.g., Stage 1 iteration 18 of the CCU model).

\subsubsection{Hitting the maximum context size}

For the stroke model, both modellers found that the context size was not large enough to generate the full model. The effects of this manifested in three ways: 

\begin{itemize}
    \item Perplexity, for all intents and purposes, ``forgot" a section of model code. For example, it could no longer output the \verb|Experiment| class code and hence could not perform the required modifications (e.g., iteration 18 of Stage 1).
    \item Incorrect code indentation at the end of the generated output (this is important because Python is whitespace sensitive; e.g., iteration 24 of stage 1).
    \item Introduction of small errors at the end of the generated text. For example, in a Python class, the `self' keyword was omitted when referring to member attributes.
\end{itemize}

We solved this problem and were able to get the correct output by switching to a new context window (i.e. a new chat prompt with no history). This meant that we were required to include code snippets (e.g., functions, classes or scripts) in the prompt to fill in the missing context.

\subsubsection{Time-consuming prompt engineering}

We used a mix of prompt types to generate the models. The time to design these prompts and their complexity varied significantly. For some difficult tasks, we were required to use one-shot or few-shot prompt engineering. An example of this was to modify code when allocating seeds across the model's random number generators (where the LLM had to both modify code and count). This meant we explicitly included Python code examples of what we wanted from the prompt.  For this specific example we found, in both stages, that a prompt needed to be split in two to achieve what we wanted with Perplexity. Informally, we tested the same prompt with OpenAI's ChatGPT and found that it could generate the desired seed allocation from our single prompt. We conclude two things from this example. The first is that if few-shot prompt engineering requires a modeller to write Python code then it is likely cost effective to simply make the changes manually and then reprompt the LLM (including the modified code).  The second is that a user may wish to try multiple AI tools when they encounter a difficulty.

\subsubsection{Changing nature of tools}

We found that the behaviour and performance of Perplexity varied during our study.  During the construction of Stage 1's CCU model, Perplexity provided sources with each prompt, meaning that retrieval-augmented generation was used to provide additional context before code generation. This behaviour changed over subsequent models and stages: for example, we found that Perplexity's sources appeared for the first two iterations of the stroke capacity planning model and then no further sources were used (displayed to the user) even when we started a new context window.  We re-tested this in early 2025 to find the functionality restored.

During each stage, we also found that the speed and performance of Perplexity's generation varied. In particular, the rate at which new code was generated would drop significantly and potentially would not fully generate the code we required. We did not include a formal investigation of this aspect in our study plan but note that this typically occurred in the afternoon in the UK.  We cannot explain this conclusively but speculate that this could have been related to the rapidly growing user base of tools like Perplexity, time difference with the US, and the prioritisation of limited GPU capacity to paying customers.  This finding may have also been compounded by the large contexts we were creating and passing to Perplexity over multiple iterations of the models.  We handled this problem by pausing the study and re-prompting the following morning.

\subsubsection{Internal reproduction, randomness and preservation}

We expected that our internal test of reproduciblity would be inherently challenging, due to the way generative AI makes use of random noise. In other words, we expected that given the same prompts, any  AI tool would generate Python code with subtle to substantial differences. As the model code in Stage 2 deviated from Stage 1 across iterations, the original prompts became less specific to the Stage 2 model implementation. As such (with the current generative AI capability) it seems unlikely that a general set of sequenced prompts could be expected to reliability regenerate a specific Python simulation model. This finding emphasises the importance of preservation of simulation model artefacts for the long term (so regeneration is unnecessary).

\subsection{Wider opportunities presented by LLM for coding Python models}

We found in all cases that our approach to recreating a version of the published models was successful. We propose the following wider opportunities for the modelling and simulation community.

\subsubsection{Reuse of prompts}

For simpler tasks, we found that some prompts were fairly generic and can work well if a user takes a ``lift and adapt approach". One example of this is generating a basic user interface to the model that we split over three iterations.  We provided the interface prompt for the stroke capacity planning model in Box \ref{box:numbered_steps}; this was a minor adaptation of the prompt used for the CCU model (see iteration 19).  These are simple prompts that begin the same and then are tailored to the specifics of the results summary for each model.  

\subsubsection{Describing models}

At the end of the study we also explored the use of generative AI to describe model code.  We used the persona pattern \citep{olea2024evaluating} from prompt engineering and asked the AI tool to take the role of an elective surgery patient.  We provided a prompt including all the CCU model code and asked for a description of potential journeys through the model.  We found that the routing through the model was described accurately.  We provide an excerpt from the response in Box \ref{box:code_description}.

\subsubsection{Documenting model code}

We informally tested the ability of LLM's to generate code documentation for stage 1 models.  Specifically we issued the prompt ``Write PEP257 compliant docstrings for all functions, classes and  methods. Provide a brief description of the purpose of the code, document parameter and return values". The generated output was promising and likely warrants further investigation. We provide an example output in Appendix \ref{app:opp}.

\begin{textbox}[h!]
\begin{textexcerpt}

\textbf{Summary of Potential Journeys}

As an elective surgery patient, my journey through the simulation model can either result in successful admission, surgery, recovery, and discharge, or in the cancellation of my surgery due to a lack of available beds. Here’s a condensed view of both paths:

\textbf{Successful Admission and Treatment:}
\begin{itemize}
    \item Arrival → Bed Request → Bed Available → Admission → Surgery and Recovery → Discharge and Cleaning
\end{itemize}

\textbf{Surgery Cancellation:}
\begin{itemize}
    \item Arrival → Bed Request → No Bed Available → Surgery Cancellation
\end{itemize}
        
\textbf{Implications for the Simulation}

These paths are critical in understanding the performance of the critical care unit. My journey as an elective surgery patient affects:

\begin{itemize}
    \item \textbf{Bed Utilization}: The time I spend in the bed contributes to overall bed utilization metrics.
    \item \textbf{Bed Occupancy}: My admission impacts bed occupancy rates.
    \item \textbf{Surgery Cancellations}: If my surgery is cancelled, it adds to the count of cancelled elective operations, providing insight into the hospital’s capacity challenges.
\end{itemize}
 
\end{textexcerpt}
\caption{Generated description of the critical care unit model (using the persona pattern for an Elective Patient).}
\label{box:code_description}
\end{textbox}

\subsection{General risks presented by LLM for coding Python models}

As we have already outlined, LLMs make mistakes when coding. For the immediate future, we caution all users that no code generated by an AI tool should be trusted at face value. Our testing and code differencing procedures were essential to identify these errors, as they could be subtle and easily missed with casual inspection. For example, a critical line of stroke patient routing logic was removed in Stage 1 iteration 3 (fixed via a prompt in iteration 11). Although verification is recommended in all DES textbooks as a hallmark of study quality, formal verification of model code is rarely reported in the DES literature \citep{ZHANG2020506}. Our findings reinforce this message and we recommend all authors who make use of Generative AI write and publish formal tests of their models alongside their model.

A secondary recommendation for users building or generating FOSS models is to make use of simple user interfaces for informal testing. We recognise that the time to test models in a real simulation study is limited. In our study we found that the generation of a basic \textit{streamlit} user interface in addition to our formal design of tests was productive. Specifically, this provided a quick way to conduct combinations of extreme value tests. As an example, this allowed us to identify an error in the logic to generate result charts from the stroke model (fixed in Stage 1 iteration 30).

\subsection{The replicability of the reported models}

Just like our two selected academic papers, the vast majority of DES healthcare models published in the academic literature take the form of a natural language description (with diagrams in some cases) and no coded model is available \citep{monks_harper_des_review}.  The ability to replicate computer models and results reported in the literature (DES and from other related methods) has long been a concern for the community \citep{mcmanus2019can,grimm2020odd,schwander_replication_2021}. We found that we were unable to fully replicate the reported results of one of the models. The primary reason for the difference we believe is straightforward: we did not have access to the empirical distributions used in the original study and we were forced to make an assumption (albeit one alluded to as an alternative in the original paper).  We believe that the design of the recreated model is still faithful to the original work and if the original parameters were reported, we hypothesise that replication of the experiments we tested would have been successful. While this may seem a minor discrepancy, we note that we cannot fully verify the model is a recreation of the original work without this data. To improve replication success, our recommendation is therefore that authors provide full verification data to support narrative descriptions of their models. Formal guidance on documenting models exists elswhere \citep{ZHANG2020506, stress_guidelines}. A specific example reporting continuous and discrete empirical distribution functions can be seen in the online supplementary material that accompanies \cite{Lahr2013}.

\subsection{Limitations and further work}

A limitation of our study is that we only investigated coding models in Python and \textit{SimPy} and further limited this to two simulation models. This narrow focus allowed us to increase the complexity of simulation coding tasks in a way that was above prior studies and to include an internal repetition to investigate reproducibility.  A natural extension to this work would be to investigate generation of models in the other popular FOSS language for DES: R and RSimmer. 

The study was also limited to two modellers that tested an iterative process to model coding. Although the prompt engineering process we followed was successful, other modellers may not wish to design their models in such an iterative manner. We did not formally test an all-in-one prompt approach but, given the challenges in catching hallucinations and mistakes, we would caution against such a strategy in practice.

Finally, we limited our formal investigation to the free tier of a single AI tool - Perplexity.AI's standard model - during a period in 2024 and cannot evidence generalisability to other popular tools.  Our informal testing leads us to speculate that our findings should translate to paid tiers of services such as Perplexity that offer access to larger models or recent models with reasoning capabilities (e.g., GPT-4.5, Claude 3.7 or DeepSeek-R1).  We expect these larger models to be at least as good as the tool we used and still advise rigorous testing of any simulation code they generate.

Looking to the future, we note that the growing presence of AI-generated content on the internet is of increasing concern \citep{xing2025caveats}. The potential challenges it poses, especially concerning the training of LLMs on AI-generated content, raise concerns about ``model collapse", a degenerative process where models trained predominantly on synthetic data experience performance degradation. This occurs because models recursively learn from outputs that may contain errors or biases, leading to a decline in quality over successive training iterations \citep{shumailov2024ai,wenger2024ai}. Future work for the modelling and simulation community may have to consider the need for careful curation of quality simulation training datasets to maintain integrity and performance.

\section{Conclusions}

This study explored the feasibility of using generative AI to recreate published DES models in healthcare. Our findings suggest that it is possible to generate DES models in FOSS using engineered prompts derived from narrative descriptions, as demonstrated by our recreation of two published healthcare models in Python. These models passed human-developed verification tests and incorporated simple user interfaces. This study also revealed notable challenges associated with generative AI, including issues with prompt engineering, code generation, and model testing. While we successfully generated, validated, and reproduced these models, our experience underscores the importance of iterative refinement, systematic approaches to model differencing and code testing, and domain expertise in DES modeling, FOSS, and healthcare. These findings provide valuable insight into the promise and limitations of generative AI for healthcare modelling, but further research is needed to fully understand its broader applicability and scalability.

\section*{Funding}

This work is supported by the NIHR Applied Research Collaboration South West Peninsula. The views expressed in this publication are those of the author(s) and not necessarily those of the NIHR or the Department of Health and Social Care. This work was supported by the Medical Research Council [MR/Z503915/1].

\section*{Data availability statement}

All prompts, code and other artefacts used in this study have been archived on Zenodo. Links to artefacts are provided in Table \ref{tab:artefacts}.

\section*{Statement of contribution}

T.Monks and A.Harper conceived of the presented idea. T.Monks and A.Harper developed the methodology and performed the analysis. A.Heather tested, verified, and modified the code and research compendium. All authors discussed the results and contributed to the final manuscript.

\bibliography{refs}

\pagebreak
\begin{appendices}

\section{Sample \textit{SimPy} Code}
\label{app:simpy_example}

For those unfamiliar with \textit{SimPy}, a simple simulation model of an urgent care call centre is presented below. This is based on introductory tutorial material published elsewhere \citep{monksharper_simpy_tutorial}. In the interest of space, we have removed docstrings and comments from the code.  The model consists of three parts. First is the patient generator function \verb|arrivals_generator| that generates inter-arrival times following an exponential distribution. Second, a \verb|service| function where patient processes request call operator resources and when available samples a call duration from a triangular distribution.  A simulated trace is provided as the model runs.  The final part of the code is a script to run the model. It creates a \textit{SimPy environment}  (that holds the DES event scheduling engine), the call operator resources, schedules the patient generator function, and starts the simulation run for a user-specified run length.

\begin{lstlisting}[language=python, caption=Example \textit{SimPy} code., label=code:example_simpy]
import simpy
import numpy as np
import itertools

def service(identifier, operators, env):
    start_wait = env.now

    with operators.request() as req:
        yield req

        waiting_time = env.now - start_wait
        print(f'operator answered call {identifier} at ' \
              + f'{env.now:.3f}')

        call_duration = np.random.triangular(left=5.0, mode=7.0,
                                             right=10.0)
        
        yield env.timeout(call_duration)

        print(f'call {identifier} ended {env.now:.3f}; ' \
              + f'waiting time was {waiting_time:.3f}')


def arrivals_generator(env, operators):
    for caller_count in itertools.count(start=1):

        inter_arrival_time = np.random.exponential(60/100)
        yield env.timeout(inter_arrival_time)

        print(f'call arrives at: {env.now:.3f}')

        env.process(service(caller_count, operators, env))


if __name__ == '__main__':
    RUN_LENGTH = 100
    N_OPERATORS = 13
    
    env = simpy.Environment()
    operators = simpy.Resource(env, capacity=N_OPERATORS)
    
    env.process(arrivals_generator(env, operators))
    env.run(until=RUN_LENGTH)
    print(f'end of run. simulation clock time = {env.now}')
\end{lstlisting}

\pagebreak
\section{Model interfaces}
\label{app:interfaces}

\setcounter{figure}{0}
\renewcommand{\thefigure}{B\arabic{figure}}

\begin{figure}[H]
    \centering
    \begin{subfigure}{\linewidth}
        \centering
        \includegraphics[width=0.65\linewidth]{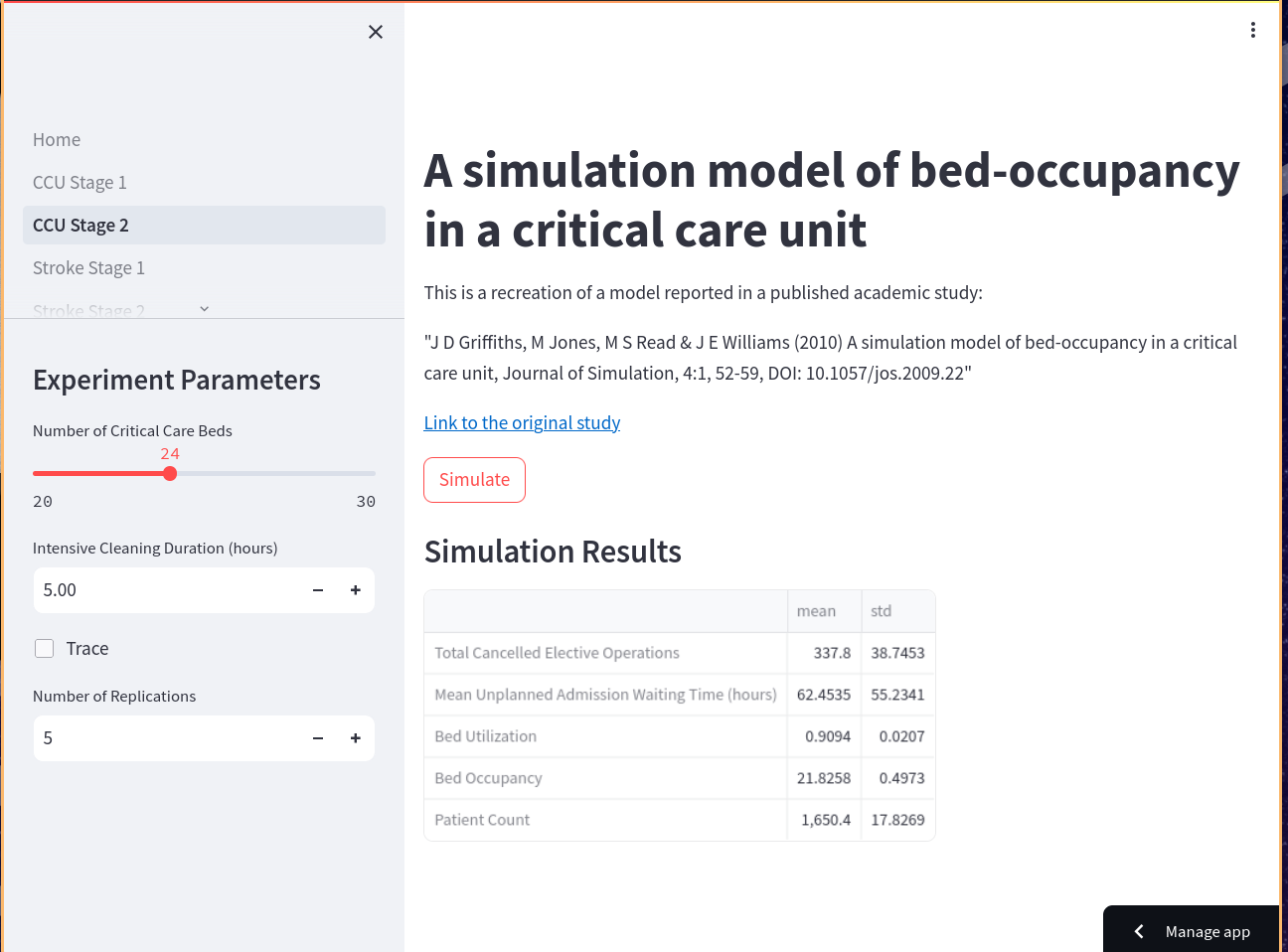}
        \caption{Generated streamlit interface for Critical Care Unit.}
        \label{fig:ccu_interface_stage2}
    \end{subfigure}
    
    \vspace{1cm}  
    
    \begin{subfigure}{\linewidth}
        \centering
        \includegraphics[width=0.65\linewidth]{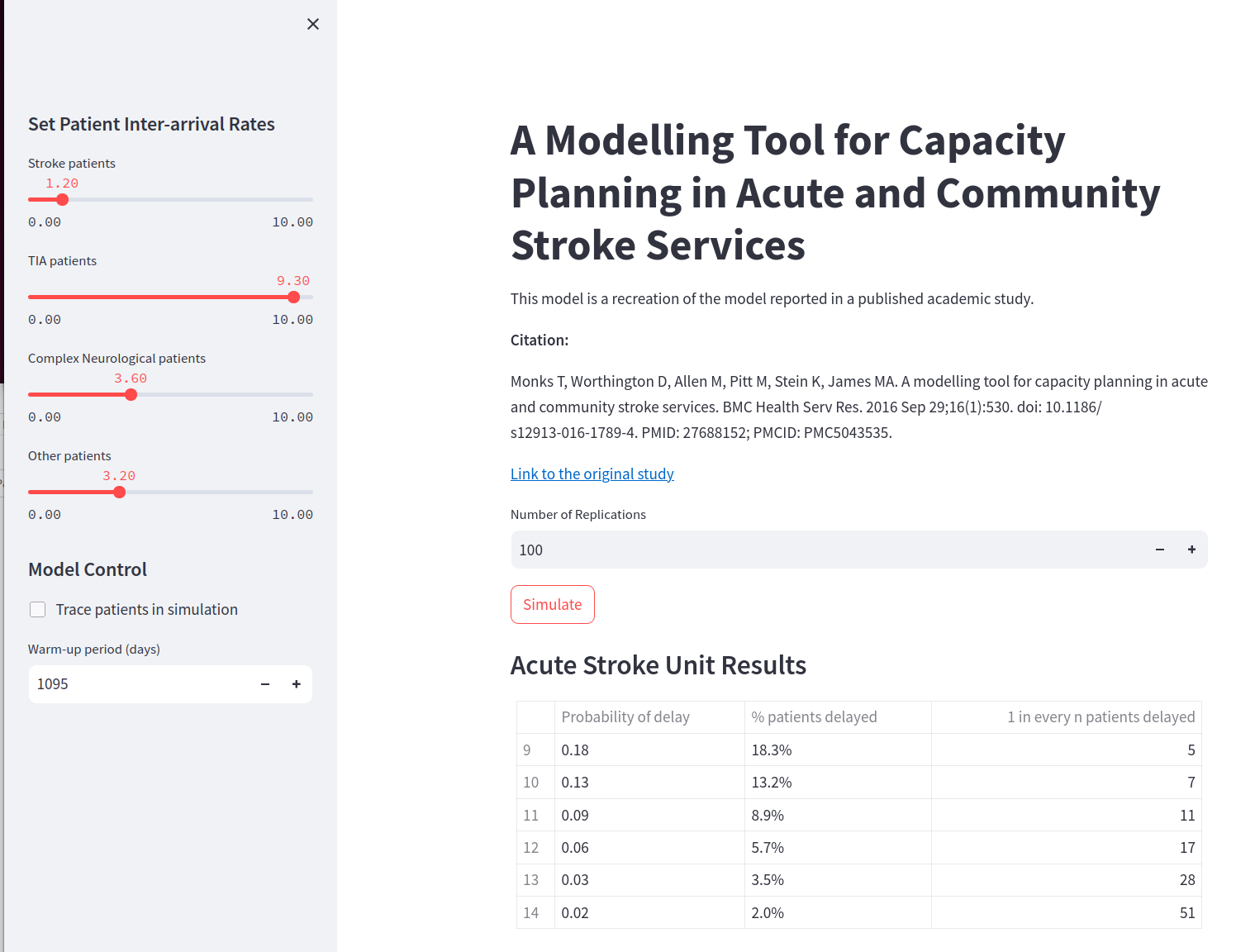}
        \caption{Generated interface fo the stroke capacity planning model.}
        \label{fig:stroke_interface_stage2}
    \end{subfigure}
    
    \caption{Model interfaces from Stage 2.}
    \label{fig:combined_interfaces_stage2}
\end{figure}

\pagebreak
\section{Prompt Comparison}
\label{app:ccu_results}

\setcounter{table}{0}
\renewcommand{\thetable}{C\arabic{table}}

\setcounter{figure}{0}
\renewcommand{\thefigure}{C\arabic{figure}}

\begin{table}[H]
\caption{The number of prompts given to the LLM at each iteration of the CCU model.}
\begin{tabular}{llrrr}
\textbf{Iteration} & \textbf{Added functionality} & \textbf{Stage 1} & \textbf{Stage 2} & \textbf{Difference} \\ \hline
1 & Unplanned arrivals & 1 & 4 & 3 \\
2 & Add treatment & 1 & 3 & 2 \\
3 & Elective patients & 2 & 2 & 0 \\
4 & Organise input parameters & 2 & 2 & 0 \\
5 & Add a warm-up period & 1 & 1 & 0 \\
6 & Elective cancellations (KPI) & 1 & 1 & 0 \\
7 & Bed utilisation (KPI) & 2 & 2 & 0 \\
8 & Waiting time (KPI) & 1 & 2 & 1 \\
9 & Bed occupancy (KPI) & 1 & 3 & 2 \\
10 & Patient count (KPI) & 1 & 3 & 2 \\
11 & Multiple replications (1) & 1 & 1 & 0 \\
12 & Multiple replications (2) & 1 & 1 & 0 \\
13 & Multiple replications (3) & 1 & 1 & 0 \\
14 & Summarise results & 1 & 1 & 0 \\
15 & Common random numbers (1) & 1 & 1 & 0 \\
16 & Common random numbers (2) & 1 & 1 & 0 \\
17 & Common random numbers (3) & 1 & 2 & 1 \\
18 & Common random numbers (4) & 1 & 1 & 0 \\
19 & Batching experiments & 1 & 1 & 0 \\
20 & Streamlit interface (1) & 1 & 1 & 0 \\
21 & Streamlit interface (2) & 1 & 1 & 0 \\
22 & Streamlit interface (3) & 1 & 1 & 0 \\
23 & Bug fix & 1 & 0 & -1 \\ \hline
& \textbf{Totals} & \textbf{26} & \textbf{36} & \textbf{10} \\ \hline
\end{tabular}
\label{app:ccu_prompt_numbers}

\end{table}

\begin{table}[H]
\caption{The number of prompts given to the LLM at each iteration of the stroke capacity planning model).}
\begin{tabular}{lp{6cm}rrr}
 & \textbf{Added functionality} & \textbf{Stage 1} & \textbf{Stage 2} & \textbf{Difference} \\ \hline
\textbf{Iteration} &  &  &  &  \\
1 & Acute stroke unit (ASU) arrivals & 1 & 3 & 2 \\
2 & Sample post stroke unit destination & 1 & 1 & 0 \\
3 & Acute stroke unit length of stay (1) & 1 & 2 & 1 \\
4 & Acute stroke unit length of stay (2) & 1 & 1 & 0 \\
5 & Organise parameters & 1 & 1 & 0 \\
6 & Track ASU bed occupancy & 1 & 1 & 0 \\
7 & Functionality to suppress simulation event log & 2 & 2 & 0 \\
8 & ASU results collection functionality & 1 & 2 & 1 \\
9 & ASU occupancy plot & 1 & 1 & 0 \\
10 & ASU probability of delay & 1 & 1 & 0 \\
11 & Bug fix: add back in code that was removed by LLM & 1 & 0 & -1 \\
12 & Rehab external arrivals & 1 & 1 & 0 \\
13 & Organise parameters & 2 & 2 & 0 \\
14 & Rehab unit length of stay (1) & 1 & 1 & 0 \\
15 & Organise parameters & 2 & 3 & 1 \\
16 & Rehab unit length of stay (2) & 1 & 1 & 0 \\
17 & Track rehab unit bed occupancy & 1 & 1 & 0 \\
18 & Rehab unit results collection functionality & 2 & 2 & 0 \\
19 & Link ASU and Rehab models (1) & 1 & 1 & 0 \\
20 & Link ASU and Rehab models (2) & 2 & 6 & 4 \\
21 & Warm-up period & 2 & 5 & 3 \\
22 & Multiple replications (1) & 1 & 1 & 0 \\
23 & Multiple replications (2) & 1 & 2 & 1 \\
24 & Common random numbers (1) & 1 & 1 & 0 \\
25 & Common random numbers (2) & 1 & 1 & 0 \\
26 & Common random numbers (3) & 2 & 5 & 3 \\
27 & Common random numbers (4) & 2 & 2 & 0 \\
28 & Model interface (1) & 1 & 1 & 0 \\
29 & Model interface (2) & 1 & 2 & 1 \\
30 & Model interface (3) & 1 & 1 & 0 \\
31 & Bug fix & 3 & 3 & 0 \\ \hline
 & \textbf{Totals} & \textbf{41} & \textbf{57} & \textbf{16} \\ \hline
\end{tabular}
\label{app:stroke_prompt_numbers}
\end{table}

\begin{figure}[H]
    \centering
    \includegraphics[width=0.75\linewidth]{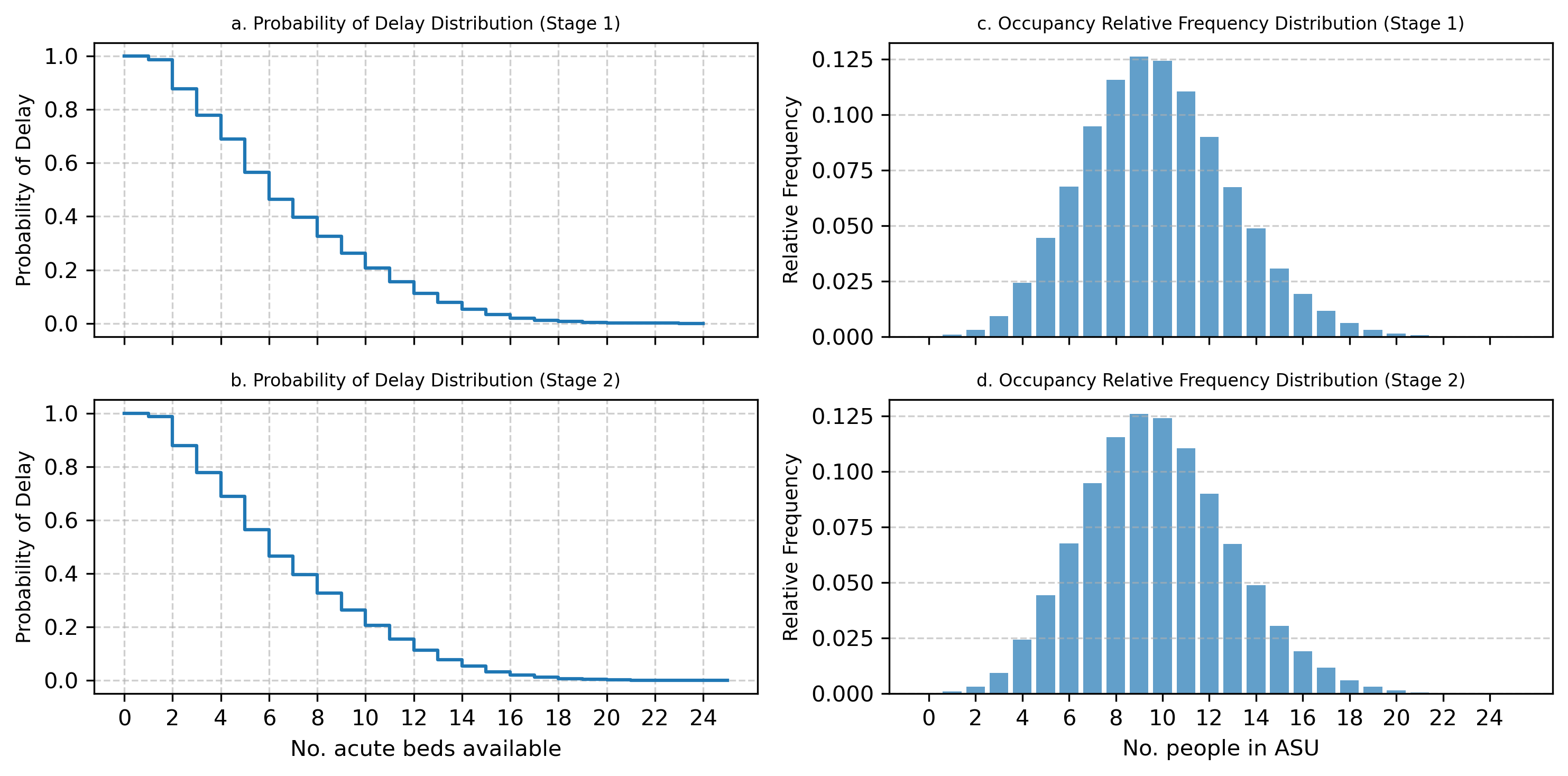}
    \caption{Rehabilitation unit outputs: comparison of stage 1 and stage 2 models}
    \label{fig:stroke_comparison_rehab}
\end{figure}

\pagebreak
\section{Wider opportunities}
\label{app:opp}

\begin{textbox}[h!]
\begin{lstlisting}[language=python, caption=Example docstrings generated by Perplexity., label=code:docs]
class AcuteStrokeUnit:
    """A class representing the Acute Stroke Unit in the simulation.

    This class models the arrival and acute treatment of stroke patients, tracking occupancy
    and simulating the duration of stay based on treatment type.

    Attributes:
    - env (simpy.Environment): The simulation environment.
    - experiment (Experiment): The experiment configuration.
    - rehab_unit (RehabilitationUnit): The connected Rehabilitation Unit.

    Methods:
    - stroke_acute_treatment, tia_acute_treatment, etc.: Simulate the treatment process.
    - stroke_patient_generator, tia_patient_generator, etc.: Generate patients over time.
    """

    def __init__(self, env, experiment, rehab_unit):
        """Initialize the Acute Stroke Unit with the simulation environment and parameters.

        Parameters:
        - env (simpy.Environment): The simulation environment.
        - experiment (Experiment): The experiment configuration.
        - rehab_unit (RehabilitationUnit): The connected Rehabilitation Unit.
        """
        self.env = env
        self.experiment = experiment
        self.rehab_unit = rehab_unit
        self.patient_count = 0
        self.occupancy = 0
\end{lstlisting}
\label{fig:documentation_example}
\end{textbox}

\end{appendices}

\end{document}